\documentclass[journal]{IEEEtran}
\makeatletter
\long\def\@makecaption#1#2{\ifx\@captype\@IEEEtablestring%
	\footnotesize\begin{center}{\normalfont\footnotesize #1}\\
		{\normalfont\footnotesize\scshape #2}\end{center}%
	\@IEEEtablecaptionsepspace
	\else
	\@IEEEfigurecaptionsepspace
	\setbox\@tempboxa\hbox{\normalfont\footnotesize {#1.}~~ #2}%
	\ifdim \wd\@tempboxa >\hsize%
	\setbox\@tempboxa\hbox{\normalfont\footnotesize {#1.}~~ }%
	\parbox[t]{\hsize}{\normalfont\footnotesize \noindent\unhbox\@tempboxa#2}%
	\else
	\hbox to\hsize{\normalfont\footnotesize\hfil\box\@tempboxa\hfil}\fi\fi}
\makeatother

\usepackage{graphicx}
\usepackage{float}
\usepackage{wrapfig}
\usepackage{algpseudocode}
\usepackage{multirow}
\usepackage{amsfonts}
\usepackage{mathrsfs}

\usepackage{amssymb,latexsym}
\usepackage{commath}
\usepackage{caption}
\usepackage{subcaption}

\usepackage{booktabs}
\usepackage{mathtools}

\usepackage{todonotes}
\usepackage{soul}
\usepackage[ruled,vlined]{algorithm2e}
\usepackage{multicol}
\usepackage{amsbsy}

\usepackage[noadjust]{cite}

\ifCLASSINFOpdf

\else

\fi

\usepackage[T1]{fontenc} % optional enhanced font encoding
\usepackage{amsmath}

\usepackage[cmintegrals]{newtxmath}
\usepackage{bm}

\usepackage{array}

\usepackage{url}

\ifCLASSINFOpdf
\else
\fi
\providecommand{\hypersetup}[1]{\relax}

\hypersetup{pdftitle={Bare Demo of IEEE\_lsens.cls for IEEE Sensors Letters},%<!CHANGE!
pdfsubject={Typesetting},%<!CHANGE!
pdfauthor={Michael D. Shell},%<!CHANGE!
pdfkeywords={Class, IEEE, IEEE\_lsens, IEEE Sensors Letters, LaTeX, Typesetting, TeX}}%<^!CHANGE!}

\usepackage{color}
\newcommand{\vect}[1]{\boldsymbol{#1}}
\graphicspath{{figs/}}

\begin{document}

\title{Physics-constrained Deep Learning for Robust Inverse ECG Modeling}

\author{Jianxin Xie and Bing Yao$^*$ 
	
\thanks{
	Jianxin Xie and Bing Yao ($^*$corresponding author: bing.yao@okstate.edu) are with the School of Industrial Engineering and Management, Oklahoma State University, Stillwater, OK 74078 USA.
}}

\maketitle

\begin{abstract}
	 The rapid development in advanced sensing and imaging brings about a data-rich environment, facilitating the effective modeling, monitoring, and control of complex systems. For example, the body-sensor network captures multi-channel information pertinent to the electrical activity of the heart (i.e.,  electrocardiograms (ECG)), which enables medical scientists to monitor and detect abnormal cardiac conditions. However, the high-dimensional sensing data are generally complexly structured. Realizing the full data potential depends to a great extent on advanced analytical and predictive methods. This paper presents a physics-constrained deep learning (P-DL) framework for robust inverse ECG modeling. This method integrates the physics law of the cardiac electrical wave propagation with the advanced deep learning infrastructure to solve the inverse ECG problem and predict the spatiotemporal electrodynamics in the heart from the electric potentials measured by the body-surface sensor network. Experimental results show that the proposed P-DL method significantly outperforms existing methods that are commonly used in current practice.
\end{abstract}

\def\abstractname{Note to Practitioners}
\begin{abstract}
	This article is motivated by the remarkably increasing applications of advanced medical sensing and imaging technique for data-driven disease diagnosis and treatment planning. For instance, body surface potential mapping (BSPM) can be non-invasively acquired to delineate the spatiotemporal potential distribution on the body surface, enabling medical scientists to infer the electrical behavior of the heart (i.e., heart surface potential (HSP)). However, the reconstruction of HSP from BSPM is highly sensitive to measurement noise and model uncertainty. This paper presents a novel physics-constrained deep learning (P-DL) framework by encoding the physics-based principles into deep-learning for robust HSP prediction. Experimental results demonstrate the extraordinary performance of the proposed P-DL model in handling measurement noise and other uncertainty factors in inverse ECG modeling.
	
\end{abstract}


\section{Introduction}
    The development of cutting-edge medical sensing and imaging technique greatly benefits the modeling of complex dynamic systems and further facilitates effective system monitoring and optimization for smart health. For instance, electrocardiography (ECG) is a widely-used diagnostic tool to investigate and detect abnormal heart conditions. It monitors electrical signals on the torso surface initiated by the electrical depolarization and repolarization of the heart chambers, establishing a spatiotemporal pattern of the cardiac electrical activity \cite{oster1998electrocardiographic}. Effective ECG monitoring systems enable medical scientists to non-invasively access the electrical behavior of the heart by simply measuring the electrical potentials on the body surface. Twelve-lead ECG, one of the most widely-used ECG systems, is implemented by strategically distributing four electrodes on limbs and six on the chest to harvest multi-angle views of the spatiotemporal cardiac electrodynamics \cite{yang2013spatiotemporal}. However, traditional ECG systems (e.g., 12-lead ECG, 3-lead VCG) only employ a limited number of ECG sensors on the thorax and thus the resulted cardiac electrical distribution is with low spatial resolution. Accurate diagnosis of heart diseases such as acute ischemia calls upon the high-resolution ECG mapping \cite{herring2006ecg} that is capable of capturing a comprehensive 3D view of the cardiac electrical signals. Thus, body surface potential mapping (BSPM) has been developed by deploying an increased number (e.g., 120) of ECG electrodes to record the high-resolution spatiotemporal cardiac electrodynamics projected on the body surface \cite{bond2009xml, lacroix1991evaluation, rudy1999noninvasive,chen2015sparse,zhu2018optimal}.

    Rapid advances in sensing and imaging bring abundant data and provide an unprecedented opportunity for data-driven disease diagnosis and optimal treatment planning \cite{ piri2017data,lin2018selective, si2017sequential,yao2017characterizing,iquebal2020consistent,9339957}. Realizing the full potential of the sensing and imaging data depends greatly on the development of innovative analytical models. In the human heart system, the ECG signal (or BSPM) describes the spatiotemporal distribution of electric potentials on the body surface, which reflects electrical behaviors of the source, i.e., epicardial potentials. When projecting from the heart to the body surface, the spatiotemporal pattern of the cardiac electrodynamics becomes diminished and blurred, which defies the accurate characterization of local abnormal regions in the heart. Precision cardiology \cite{corral2020digital,trayanova2019genetics,yao2021spatiotemporal} calls for advanced analytical models that can effectively handle the spatiotemporal data structure for robust prediction of heart-surface electrical signals (i.e., heart-surface potentials (HSP)) from body-surface sensor measurements (i.e., BSPMs). This is also called the inverse ECG problem \cite{ rudy1999noninvasive,oster1997noninvasive,gulrajani1998forward,yao2016mesh, yao2020spatiotemporal}.
    
    Inverse ECG problem is commonly recognized as ill-conditioned, because a small input noise may cause highly unstable prediction. Great efforts have been made to achieve robust inverse solutions by adding proper spatial or/and temporal regularization \cite{brooks1999inverse, greensite2003temporal, wang2009physiological, yu2015temporal,karoui2018evaluation}. However, most existing regularization methods (e.g., Tikhonov, L1-norm, or spatiotemporal regularization) focus on exploiting the data structure (i.e., data-driven regularization). Little has been done to integrate the well-established physics model of cardiac electrodynamics (i.e., physics-driven regularization) with advanced machine learning approaches for robust inverse ECG modeling. In order to further increase the estimation accuracy and robustness, it is imperative to incorporate the physics prior knowledge that theoretically describes the cardiac electrical wave propagation into the transformation from BSPM data to heart-surface signals. This paper proposes a novel physics-constrained deep learning (P-DL) framework to solve the inverse ECG problem and achieve a robust estimation of HSP from noisy BSPM data. Specifically, our contributions in the present investigation are as follows:
    	
    $\bullet$	We propose to integrate the physics prior knowledge of cardiac electrodynamics with advanced deep learning (DL) infrastructures to regularize the inverse ECG solution. The loss function of the neural network is delicately designed to not only satisfactorily match the sensor measurements (i.e., data-driven loss) but also respect the underlying physics principles (i.e., physics-based loss).
    
    $\bullet$	The effect of the physics-driven regularization (i.e., the physics-based loss) on the inverse solution will be controlled by introducing a regularization parameter, the value of which will impact the prediction performance. We define a balance metric to effectively evaluate the prediction performance of the proposed P-DL framework given different values of the regularization parameter. We further propose to model the balance metric with Gaussian Process (GP) regression, and effectively search for the optimal regularization parameter using the GP upper-confidence-bound (GP-UCB) algorithm.
    
    We validate and evaluate the performance of the proposed P-DL framework to reconstruct the HSP from BSPM data in a 3D torso-heart geometry. Numerical experiments demonstrate the superior performance of our P-DL approach compared with existing regularization methods, i.e., Tikhonov zero-order, Tikhonov first-order, spatiotemporal regularization, and the physiological-model-constrained Kalman Filter method.

    The remainder of this paper is organized as follows: Section II presents the literature review of the inverse ECG problem. Section III introduces the proposed P-DL framework. Section IV shows the numerical experiments of the P-DL method to solve the inverse ECG problem in a 3D torso-heart geometry. Section V discusses the future work and advantages of the proposed P-DL method over existing regularization methods in solving the inverse ECG problem. Section VI concludes the present investigation.

%%%%%%%%%%%%%%%%%%%%%%%%%%%%%%%%%%%%%%%%%%%%%%%%%%%%%%%%%%%%%%%%%%%%%%%%%%%%%%%
\section{Literature Review}

    \subsection{Inverse ECG Modeling and Regularization}
   	 The goal of inverse ECG modeling is to enable non-invasive reconstruction of cardiac electrical activity from body-surface sensor measurements for optimal diagnosis and characterization of heart disease. It has been widely recognized that inverse ECG problem is ill-posed, the solution of which is sensitive to model uncertainty and measurement noise. Specifically, the relationship between  BSPM $ \vect{y}( \vect{s}, t) $ with HSP $ \vect{u}(\vect{s}, t) $ can be expressed as $\vect{y}(\vect{s}, t)=\vect{Ru}(\vect{s}, t)+\epsilon$, where $ \vect{s} $ and $t$ denotes the spatial and temporal coordinates respectively, and $ \vect{R} $ is a transfer matrix solved by Divergence Theorem and Green's Second Identity \cite{yao2020spatiotemporal,barr1977relating}. However, the rank-deficiency of transfer matrix $ \vect{R} $ (i.e., $ \textrm{rank}(\vect{R})<\min \{ \dim(\vect{u}), \dim(\vect{y}) \} $) and its large condition number (i.e., $ cond(\vect{R}=\|\vect{R}\|\|\vect{R}^{-1}\| $) make the estimated HSP $ \vect{\hat{u}}(\vect{s}, t) $ very sensitive to the BSPM data $ \vect{y}(\vect{s}, t) $. Small fluctuation in BSPMs, i.e., $\Delta\vect{y}$, will cause a significant variation $\Delta\vect{\hat{u}}$ in $\vect{\hat{u}}$ (i.e., $ \Delta\vect{\hat{u}}/\vect{\hat{u}}\backsimeq cond(\vect{R})\Delta\vect{y}/\vect{y} $). Therefore, regularizing the inverse ECG model for robust solution is crucially important for understanding the mechanism of the cardiac physiology and pathology, further facilitating the accurate disease diagnosis and optimal therapeutic design.
   		
   	In existing literature, remarkable progress has been made in developing statistical regularization algorithms by exploiting the spatiotemporal data structure to improve the robustness of inverse ECG solution. Among them, Tikhonov method (including zero-order and first-order) is one of the most popular regularization approaches. It minimizes the mean squared error (i.e., residual) from body-heart transformation and penalizes the $ L_2 $ norm of inverse HSP solution \cite{rudy1999noninvasive, dawoud2008using, wang2011finite,throne2000fusion} to suppress the unreliable component and improve the overall smoothness of the solution. The objective function for Tikhonov regularization is formulated in Eq.(\ref{Eq:tikh}) 
 	\begin{eqnarray}
 		\label{Eq:tikh}
 		\min_{\vect{u}(\vect{s},t)} \{ \norm{ \vect{y}(\vect{s},t)-\vect{Ru}(\vect{s},t)}_2^2+\lambda_{Tikh}^2 \norm{ \Gamma \vect{u}(\vect{s},t)}_2^2 \}        
 	\end{eqnarray}         
   	where $\norm{\cdot}_2 $ indicates the $ L_2 $ norm, $ \lambda_{Tikh} $ is a regularization coefficient, and $ \Gamma $ denotes the operator constraining HSP solution $ \vect{u}(\vect{s},t) $, which is an identity matrix for zero-order regularization and a spatial gradient operator for first-order regularization to increase the spatial smoothness of the inverse solution. Another widely-used regularization technique is the $ L_1 $ norm-based method including zero-order \cite{wang2011finite} and first-order \cite{ghosh2009application, shou2010epicardial}, which penalizes $ L_1 $ norm of inverse HSP solution to encourage the model sparsity and regularity. However, both Tikhonov and $ L_1 $ regularization methods only consider the spatial data structures in inverse ECG modeling, but do not effectively incorporate the temporal evolving patterns of the cardiac electrodynamics.
   		
   	A variety of spatiotemporal regularization methods have been developed to fuse the space-time correlations into the inverse ECG problem for robust estimation of HSP \cite{ brooks1999inverse, greensite2003temporal, wang2009physiological, yu2015temporal,yao2020spatiotemporal}. For example, Messnarz \textit{et al.} \cite{messnarz2004new} reconstructed the potential distribution on the heart surface by defining a symmetric matrix of the spatial gradient to account for the spatial correlation and assuming that cardiac electric potentials are non-decreasing during the depolarization phase to address the temporal correlation. Yao \textit{et al}. \cite{yao2016physics} proposed a spatiotemporal regularization algorithm by regularizing both the spatial smoothness and increasing temporal robustness in inverse ECG modeling, whose objective function is given by Eq. (\ref{Eq:STRE}).
   		\begin{eqnarray}
   		\label{Eq:STRE} 
   		\begin{split}
   		\min_{\vect{u}(\vect{s},t)} \sum_t \Bigl\{ \ \norm{ \vect{y}(\vect{s},t)-\vect{Ru}(\vect{s},t)}_2^2+\lambda_s^2 \norm{ \Gamma \vect{u}(\vect{s},t)}_2^2 \\ 
   		+ \lambda_t^2 \sum_{\tau=t-\omega/2}^{\tau=t+\omega/2}\norm{\vect{u}(\vect{s},t)-\vect{u}(\vect{s},\tau)}_2^2 \Bigr\} 
   		\end{split}       
   		\end{eqnarray}  
   	where $\lambda_s$ and $\lambda_t$ denote the spatial and temporal regularization parameters respectively, and $\omega$ is a time window selected to incorporate the temporal correlation.
   	However, most existing spatiotemporal regularization methods mainly focus on exploiting the data structure to capture the statistical correlations between adjacent spatial locations and neighboring time points, and do not incorporate the physics principle that governs the HSP evolving dynamics into regularization for robust inverse ECG modeling.
   	 %However, most existing spatiotemporal regularization methods mainly focus on capturing the statistical correlations between adjacent spatial locations and neighboring time points, and do not incorporate the real physics principle that governs the HSP evolving dynamics over time. 
   	 Note that Wang \textit{et al}. \cite{wang2009physiological} proposed a physiological-model-constrained Kalman filter (P-KF) approach to solve the inverse ECG problem. They formulated the cardiac system into a high-dimensional stochastic state-space model and estimate the cardiac electric potential with the Kalman filter method. However, this method involves large matrices operation and may suffer from the initialization problem \cite{wang2009physiological,zhao2020trial}. Little has been done to develop innovative machine learning approaches that can incorporate the underlying physics law for robust and effective inverse ECG modeling.

    \subsection{Deep Neural Networks}
    
    Deep Neural Network (DNN) or Deep Learning (DL) has been proved to be a powerful tool as a universal function approximator that is capable of handling a variety of problems with strong nonlinearity and high dimensionality \cite{lecun2015deep}. Different advanced DL structures such as convolutional neural networks (CNNs) and recurrent neural networks (RNNs) have been developed to extract efficient features from high-dimensional data for predictive modeling \cite{yu2020robotic, wang2019octreenet, park2016online, zheng2020anatomically, wang2021generic, zheng2019cross}. However, little has been done to utilize the DNN to solve the inverse ECG problem and predict the HSP. This is due to the fact that traditional DNN algorithms require an extensive amount of data for modeling complex physical or biological systems. Such data are generally sparse and costly to acquire, which makes traditional DNNs lack robustness and fail to converge. For example, the intracardiac electrical signals are often measured by placing electrodes within the heart via cardiac catheters, which are of low spatial resolution, expensive to acquire, and incur discomforts to patients \cite{schulman1999effect}. 
    
    To cope with the problem of limited data in complex physical and biological systems, recent techniques have been developed to incorporate the underlying system physics into data-driven predictive modeling. For example, Wang et al. \cite{wang2021knowledge} proposed to constrain the GP modeling with the mechanistic model of proliferation invasion describing the dynamics of tumor growth for accurate prediction of the spatial distribution of tumor cell density. Raissi et al. \cite{raissi2017physicsII} innovatively developed a physics-informed neural network (PINN) framework that embeds the underlying physical law of nonlinear high-dimensional systems in the form of partial differential equations (PDE) into the DNN. The PINN algorithm has been further adapted to solve various problems including simulating incompressible fluid flow by Navier–Stokes equation \cite{jin2020nsfnets}, elastodynamic problem \cite{rao2020physics}, heat source layout optimization \cite{chen2020heat}, material designing \cite{liu2019multi}, physical law discovery from scarce and noisy data \cite{berg2019data, both2020deepmod}, cardiac activation by solving Eikonal equation \cite{sahli2020physics}, and cardiovascular flows modeling \cite{kissas2020machine}.
    
    As opposed to the traditional DNN modeling, where the mechanism underlying the physics system is generally unknown and has to be learned from the observed data, there exists a vast amount of prior knowledge about the cardiac dynamic system which is characterized by the physics-based PDEs. However, very little has been done to leverage the advanced deep learning infrastructure and incorporate the physics-based cardiac model for robust inverse ECG modeling. In this work, we propose a P-DL framework to encode physics-based principles into the DNN to robustly reconstruct the dynamic potential mapping on the heart surface from BSPMs. Additionally, we propose to efficiently search for the optimal balance between the physics-based loss and the data-driven loss using GP-UCB. The proposed P-DL framework not only effectively incorporates the body-surface sensing data into ECG modeling but also respects the underlying physics law, thereby significantly increasing the robustness and accuracy of the inverse ECG solution.

%%%%%%%%%%%%%%%%%%%%%%%%%%%%%%%%%%%%%%%%%%%%%%%%%%%%%%%%%%%%%%%%%%%%%%%%%%%%%%%%%%%%
   \begin{figure*}
   	\centering
   	\includegraphics[width=5.5in]{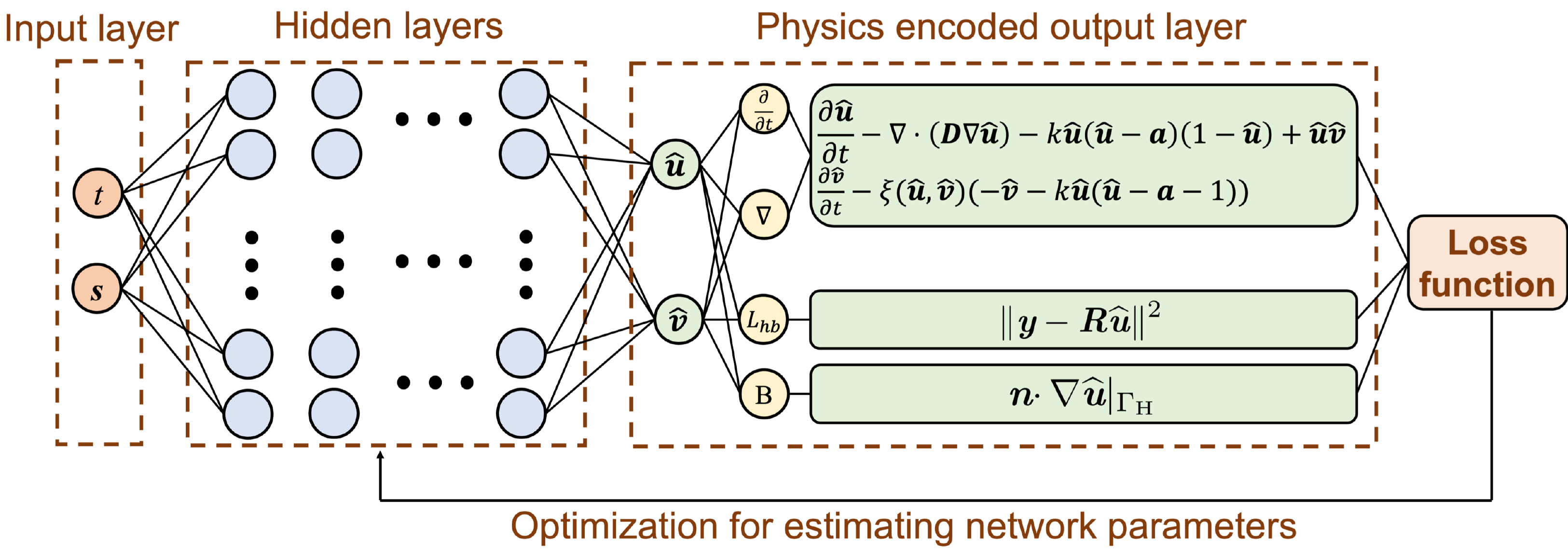}
   	%\captionsetup{justification=centering}
   	\caption{Illustration of the proposed P-DL framework for HSP prediction. Note that $ \widehat{\vect{u}} $ and $ \widehat{\vect{v}} $ are the predicted HSP and recovery current, respectively. The physical constraint of AP model and boundary condition are converted into residuals as physical loss $ \mathcal{L}_{ph} $ and added to the total loss function of the DNN for respecting the underlying physics principle. }
   	\label{Fig:structure}
   \end{figure*}

\section{Research Methodology}
    \subsection{Cardiac Electrophysiological Model}
     Different phenomenological models have been developed to study cardiac electrophysiology and delineate the electrical wave propagation in the heart \cite{sermesant2006electromechanical, camara2011inter}. Here,  without loss of generality, we utilize Aliev–Panfilov (AP) model \cite{aliev1996simple} that is widely used in various cardiac research to describe the spatiotemporal evolution of cardiac electrodynamics:
     \begin{eqnarray}
     \frac{\partial u}{\partial t}&=&\nabla \cdot (D\nabla u)+k_ru(u-a)(1-u)-uv  \label{ep:u}\\    	  
     \frac{\partial v}{\partial t}&=&\xi(u,v)(-v-k_ru(u-a-1)) \label{ep:v} \\	
     \vect{n}\cdot\nabla u|_{\Gamma_H}&=&0
     \label{ep:boundary}
     \end{eqnarray}
     where $ u $ denotes the normalized HSP, $ v $ represents the recovery current that controls the local depolarization behavior of the cardiac electric potential, and $\nabla$ denotes a spatial gradient operator on the heart geometry. Parameters $ \xi(u,v)=e_0+\mu_1v/(u+\mu_2) $ describes the coupling between $ u $ and $ v $, $ D $ is the diffusion conductivity, $ k_r $ is repolarization constant, and $ a $ controls the tissue excitability. The value of each parameter for the remainder of this research are set as $ a=0.1 $, $ D=10 $, $ k_r=8 $, $ e_0=0.002 $, $ \mu_1=\mu_2=0.3 $ from the documented literature \cite{aliev1996simple}. Note that Eq. (\ref{ep:boundary}) represents the boundary condition to guarantee there is no electric current flowing out from the heart, where $\Gamma_H$ is the heart boundary and $\vect{n}$ denotes the surface normal of $\Gamma_H$. The electrophysiological model in Eq. \eqref{ep:u}-\eqref{ep:boundary} will further be used to constrain the DNN model to achieve a robust and accurate inverse ECG solution.

    \subsection{Artificial Neural Networks}

    Deep learning has increasingly drawn significant attention due to its successful applications in a variety of fields, such as image and character recognition, natural language processing, advanced manufacturing, and computer vision \cite{xiao2020depth,wang2021multi,wang2018deep}. Feed-forward fully-connected neural network is the most fundamental architecture, which consists of an input layer, multiple hidden layers, and one output layer. The information passes layer-by-layer in one direction. The value of each neuron in a layer is calculated by the sum of the products of weights and outputs from the previous layer, which is then activated by a specific activation function. The relationship between two neighbor layers, layer-$ (n-1) $ and layer-$ (n) $, is generally described by the following equation:
    \begin{equation}
    \vect{x}_{n}=\sigma(\vect{b}_n+\vect{W}_n\vect{x}_{n-1}), 1<n\leq K
    \end{equation}
    where $\sigma(\cdot)$ represents the activation function, $ K $ is the total number of layers, $\vect{W}_n$ and $ \vect{b}_n $ are the weight matrix and bias vector for layer-$ (n) $ respectively, and $ \vect{x}_{n}$ and $ \vect{x}_{n-1}$ denote the outputs of the $ (n-1) $-th and $ (n) $-th layers, respectively. 
    
    Our objective is to solve the inverse ECG problem for the spatiotemporal HSP $ \vect{u}(\vect{s}, t) $ through the DNN modeling. The selection of optimal neural network structure will be introduced in \ref{Subsec:NN}. The hyperbolic tangent function is chosen as an activation function due to its great training performance in multi-layer neural networks. Here, we implement the basic neural network structure with a tailored loss function that is constrained by the physics law (see Eq. \eqref{ep:u}-\eqref{ep:boundary}) to solve the inverse ECG problem, and eventually reconstruct the electrical signals on the heart surface. The detailed construction of the P-DL framework will be presented in the following subsection.

    \subsection{P-DL approach}
    \label{sec:p-dl}

    	Fig. \ref{Fig:structure} shows the proposed P-DL framework. The solution of the inverse ECG problem is parameterized by a DNN that is trained to not only satisfy the physics law but also meet the data-driven constraint incurred by the body-surface sensor measurements (i.e., BSPM). Specifically, we estimate the spatiotemporal mapping of HSP as
    	\begin{eqnarray}
    	[\vect{s},t]\xRightarrow{\mathcal{N}(\vect{s},t;\vect{\theta}_{NN})}[u(\vect{s},t),v(\vect{s},t)]
    	\end{eqnarray}
    	where $\mathcal{N}(\vect{s},t;\vect{\theta}_{NN})$ defines the DNN model, $\vect{s}$ and $t$ denotes the spatial and temporal coordinates respectively, and $\vect{\theta}_{NN}$ represents the DNN hyper-parameters. Note that the DNN, i.e., $\mathcal{N}(\vect{s},t;\vect{\theta}_{NN})$, is constructed with an input layer composed of the spatiotemporal coordinates $[\vect{s},t]$, the hidden layers to approximate the complex functional relationship, and the output layer with the estimation of $\hat{u}(\vect{s},t;\vect{\theta}_{NN})$ and $\hat{v}(\vect{s},t;\vect{\theta}_{NN})$. We further encode the physics law into the DNN by defining a unique loss function as
    	\begin{equation}
    	\mathcal{L}(\vect{\theta}_{NN} )= \mathcal{L}_{hb} + w \cdot \mathcal{L}_{ph}
    	\label{Eq: PDL}
    	\end{equation}
    	where $w$ is introduced to control the regularization effect imposed by the physics principles. The total loss $\mathcal{L}(\vect{\theta}_{NN} )$ consists of the following two key elements:
    	
    	(1) \textbf{Data-driven loss} $\mathcal{L}_{hb}$: BSPMs provide the evolving dynamics of electrical potential distribution $\vect{y}(\vect{s}, t)$ on the torso surface, which can be leveraged to estimate the heart-surface electrical signals. Specifically, the HSP $ \vect{u}(\vect{s}, t) $ and body surface potential (BSP) $ \vect{y}(\vect{s}, t) $ can be related by the transfer matrix $ \vect{R} $ (i.e., $\vect{y}(\vect{s}, t)=\vect{R} \vect{u}(\vect{s}, t)$), which is derived according to the divergence theorem and Green's Second Identity (see more details at \cite{yao2020spatiotemporal,barr1977relating}). The data-driven loss generated from the transformation between BSP and HSP, denoted as $\mathcal{L}_{hb}$, is thus defined as:
    	\begin{equation}
    	\mathcal{L}_{hb}=\frac{1}{N_{st}}\sum_t\sum_{\vect{s}}\|\vect{y}(\vect{s}, t)-\vect{R}\vect{\hat{u}}(\vect{s}, t)\|^2
    	\label{loss_hb}
    	\end{equation}  
    	where $N_{st}$ stands for the total number of the discretized spatiotemporal instances.
    	
    	As stated before, due to the big condition number of the matrix $\vect{R}$, the prediction of HSP lacks the accuracy and robustness if only minimizing the mean squared error (i.e., training the DNN with only $\mathcal{L}_{hb}$) without any regularization. Additional constraint needs to be imposed to regularize the DNN-based estimation for robust inverse ECG modeling. 
    	
    	(2) \textbf{Physics-based loss} $\mathcal{L}_{ph}$: In order to further increase the estimation accuracy and robustness, physics constraints are imposed over the spatiotemporal collocation points, $[\vect{s}_i,t_i]$'s, that are randomly selected within the spatial domain defined by the heart geometry and the temporal domain defined by the cardiac cycle. The physics constraints include two parts: the boundary condition and cardiac electrophysiological model (see  Eq. \eqref{ep:u}-\eqref{ep:boundary}). Specifically, we define the boundary condition-based residual as:
    	\begin{eqnarray}
    	r_{bc}(\vect{s},t;\vect{\theta}_{NN}):=\vect{n}\cdot \nabla \vect{\hat{u}}(\vect{s},t;\vect{\theta}_{NN}),~~~~ \vect{s}\in \Gamma_H
    	\end{eqnarray}
    	The boundary condition will then be achieved by encouraging $r_{bc}(\vect{s},t;\vect{\theta}_{NN})$ to be close to zero. As such, we define the boundary condition-based loss as
    	\begin{eqnarray}
    	\mathcal{L}_{bc}=\frac{1}{N_{bc}}\sum_{i=1}^{N_{bc}}(r_{bc}(\vect{s}_i,t_i;\vect{\theta}_{NN}))^2, ~~~~\vect{s}_i\in \Gamma_H
    	\end{eqnarray}    	
    	where $N_{bc}$ denotes the number of collocation points selected on the boundary $\Gamma_H$. Additionally, the electrophysiological model-based residuals are defined as
    	\begin{eqnarray}
    	r_{u}(\vect{s},t;\vect{\theta}_{NN})&:=&\frac{\partial \hat{u}}{\partial t}-\nabla \cdot (D\nabla \hat{u})-k_r\hat{u}(\hat{u}-a)(1-\hat{u}) + \hat{u}\hat{v}\nonumber\\
    	r_{v}(\vect{s},t;\vect{\theta}_{NN})&:=&\frac{\partial \hat{v}}{\partial t}-\xi(\hat{u},\hat{v})(-\hat{v}-k_r\hat{u}(\hat{u}-a-1))
    	\end{eqnarray}
    	The physiological-based constraint will be achieved by minimizing the magnitude of both $r_{u}(\vect{s},t;\vect{\theta}_{NN})$ and $r_{v}(\vect{s},t;\vect{\theta}_{NN})$. Thus, the electrophysiological model-based loss is defined as
    	\begin{eqnarray}
    	\mathcal{L}_f=\frac{1}{N_f}\sum_{i=1}^{N_{f}}((r_{u}(\vect{s}_i,t_i;\vect{\theta}_{NN}))^2+(r_{v}(\vect{s}_i,t_i;\vect{\theta}_{NN}))^2)
    	\end{eqnarray}
    	where $N_f$ denotes the total number of selected spatiotemporal collocation points to incorporate the AP model. The physics-based loss combines the effect from both the boundary condition and the electrophysiological model, i.e., $\mathcal{L}_{ph}=\mathcal{L}_{bc}+\mathcal{L}_{f}$, which is encoded to the DNN loss function to respect the underlying physics laws in inverse ECG modeling.
    	
    	In order to compute the loss function during the DNN training, it is necessary to acquire the value of partial derivatives of $u(\vect{s},t)$ and $v(\vect{s},t)$ with respect to spatial or temporal variables. This can be easily achieved through automatic differentiation, which is able to yield the exact result of derivatives without approximation error (except for the round-off error) \cite{baydin2018automatic}. Automatic differentiation currently is the most popular method to train the deep neural networks by calculating the derivatives \textit{via} breaking down the differentials in multiple paths using the chain rule (See more details in Appendix A). In this study, automatic differentiation is engaged to obtain the derivatives with respect to the spatiotemporal coordinates $ [\vect{s}, t] $ in order to encode the physical laws in the form of a partial differential equation into the DNN.

   \subsection{Bayesian Optimization for Regularization Parameter}
   \label{sec:Gaussian}
    	The performance of the proposed P-DL framework also depends on the value of the regularization parameter $w$, as shown in Eq. (\ref{Eq: PDL}). In traditional regularization methods, the regularization parameter is generally selected according to the L-curve method \cite{hansen1993use}, which is ambiguous, insecure, and requires solving the inverse ECG problem repeatedly given different parameter values. This may incur a huge computation burden for large-scale inverse ECG modeling. Here, we propose to actively search for the optimal regularization parameter through the Gaussian Process (GP)-based surrogate modeling.
    	
    	GP has been widely used in machine learning as a method for solving complex black-box optimization problems \cite{seeger2004gaussian}. Here, we propose to search for the optimal regularization parameter by GP-based Bayesian optimization. Before establishing the Bayesian optimization procedure, we first define a balance metric as the criterion to evaluate the prediction performance given the parameter $w$ as
    	\begin{equation}
    	\label{metric}
    	m(w) =\log \left[ \left( \frac{\mathcal{L}_{hb}}{\mathcal{L}_{ph}}+\frac{\mathcal{L}_{ph}}{\mathcal{L}_{hb}} \right)\cdot \mathcal{L}(\vect{\theta}_{NN}) \right]
    	\end{equation}
       The term of $\left( \frac{\mathcal{L}_{hb}}{\mathcal{L}_{ph}}+\frac{\mathcal{L}_{ph}}{\mathcal{L}_{hb}} \right)$ in Eq. (\ref{metric}) controls the balance degree between the data-driven loss $L_{hb}$ and physics-based loss $\mathcal{L}_{ph}$, which will achieve its minimum value of 2 if $\mathcal{L}_{hb}=\mathcal{L}_{ph}$ (i.e., the training result is perfectly balanced between the data-driven term and physics-regularization term).  The term of $\mathcal{L} (\theta_{NN})$ controls the overall magnitude of the total loss, minimizing which will help to reduce either the data-driven or physics-based loss or both. A combination of the two terms will serve as an effective metric to measure the prediction performance of the P-DL given different values of the regularization parameter. In other words, minimizing Eq. (\ref{metric}) will enable to not only reduce the total loss $\mathcal{L}(\theta_{NN})$  but also help to balance between the data-driven loss $\mathcal{L}_{hb}$ and physics-based loss $\mathcal{L}_{ph}$, thereby improving the performance of the P-DL for robust inverse ECG modeling. 
       
       The input of the GP modeling is the regularization parameter, i.e., $w$'s, and the output is the corresponding balance metric, i.e., $m(w)$'s. As such, we assume 
    	\begin{eqnarray}
    	m(w)=\hat{m}(w)+\delta_m, ~~~~~ \delta_m\sim\mathcal{N}(0,\sigma^2_{\delta_m})
    	\end{eqnarray}
    	where $ \delta_m $ is Gaussian white noise added for numerical stability, and $\hat{m}(w)$ is further modeled as a GP, i.e., $\hat{m}(w)\sim \mathcal{GP}(0,\mathcal{K})$, where $\mathcal{K}$ denotes the kernel function which is selected as the widely used square-exponential kernel:
    	\begin{equation}
    	\mathcal{K}(w_i, w_j) = \sigma_m^2 \cdot \exp \left( -\frac{\|w_i-w_j\| ^2}{2l_m^2} \right)
    	\end{equation}
    	where $ \sigma_m $ and $ l_m $ denote the GP hyperparameters, which can be determined by maximizing the log marginal likelihood for GP regression  \cite{rasmussen2003gaussian}. Given the collected observations $\{\vect{m}_{1:n},\vect{w}_{1:n} \}$, where $ \vect{m}_{1:n} = [m_1 \textellipsis m_n]^T $ and $ \vect{w}_{1:n} = [w_1 \textellipsis w_n]^T $, the predictive distribution of $m(w)$ given an arbitrary unevaluated regularization parameter $w_*$ is
    	\begin{eqnarray}
    	P(m(w_*) | \vect{m}_{1:n}, \vect{w}_{1:n}, w_{*}) = \mathcal{N}(\mu_n(w_*), \sigma_n^2(w_*)) 
    	\end{eqnarray}
    	The predictive mean $ \mu_n(w_*) $ and variance $ \sigma_n^2(w_*) $ is given as
    	\begin{IEEEeqnarray}{rCl}
    	\label{mean}
    	\mu_n (w_*) &=& \mathcal{K}(w_*,\vect{w}_{1:n} )(\mathcal{K}(\vect{w}_{1:n},\vect{w}_{1:n}) +\sigma_{\delta_m}^2 \mathcal{I})^{-1} \vect{m}_{1:n}\\
    	\sigma_n^2(w_*) &=& \mathcal{K}(w_*, w_* )\nonumber \\
    	&&- \mathcal{K}(w_*,\vect{w}_{1:n} )(\mathcal{K}(\vect{w}_{1:n},\vect{w}_{1:n}) +\sigma_{\delta_m}^2 \mathcal{I})^{-1} \mathcal{K}(\vect{w}_{1:n},w_*)\nonumber \\
    	\label{variance}
    	\end{IEEEeqnarray}
    	where $ \mathcal{I}$ represents an identity matrix with the same dimensionality of $\mathcal{K}(\vect{w}_{1:n},\vect{w}_{1:n})$.
    	
    	With the GP-surrogate modeling, we can evaluate the prediction performance of P-DL given an arbitrary value of $ w$ without actually solving the inverse ECG problem. As such, we can intuitively select the optimal parameter by minimizing the GP predictive mean, i.e., $w_{opt}=\arg\min_w\{\mu_n(w)\}$. However, the accuracy of the GP surrogate depends on the quality of the training data $\{\vect{m}_{1:n},\vect{w}_{1:n} \}$, which requires an effective strategy to collect the most informative data points. To find a best next query point $ w_{n+1} $ in the parameter space for updating GP, a good acquisition function should balance the trade-off between exploitation of current knowledge and exploration of uncertain parameter space, which is constructed based on GP upper-confidence-bound (GP-UCB) \cite{srinivas2012information} as
    	\begin{equation}
    	w_{n+1} = \arg\max_w \{- \mu_n (w) + \beta^{1/2} \sigma_n(w) \}
    	\label{query point}
    	\end{equation}
    	where $ \beta $ is a positive constant that balances between the global search and local optimization. The optimal value of $\beta$ depends on the context of the problem. The theoretical rule to select an appropriate value for $\beta$ is provided in Srinivas et al. \cite{srinivas2009gaussian}. Specifically, a good regret bound can be achieved with high probability, if $\beta_{i}=2 \log \left(\frac{|\mathcal{A}| i^{2} \pi^{2}}{6 \delta}\right)$, where $i$ denotes the iteration index, $\mathcal{A}$ is the input space, and $\delta$ is a parameter between 0 and 1. However, this rule does not restrict other choices for $\beta$. The authors in \cite{srinivas2009gaussian} mentioned that the GP-UCB algorithm can be improved by scaling the theoretical value for $\beta_i$ down by a factor of 5. In our work, we simplify the process for $\beta$ selection and empirically select $\beta=9$. 
    	
    	After the new point $ w_{n+1} $ is selected, the P-DL framework will be run to solve the inverse ECG problem and calculate the corresponding balance metric $m(w_{n+1})$. When the new input-output pair $ \{ w_{n+1}, m_{n+1}\} $ becomes available, the GP surrogate will be updated according to: 
    	\begin{equation}
    	\begin{bmatrix}
    	\vect{m}_{1:n}\\
    	m_{n+1}
    	\end{bmatrix}
    	\sim
    	\mathcal{N} \left( 0, \begin{bmatrix}
    	\mathcal{K}(\vect{w}_{1:n},\vect{w}_{1:n})+\sigma_{\delta_m}^2 \mathbf{I} & \mathcal{K}(\vect{w}_{1:n},w_{n+1})\\
    	\mathcal{K}(w_{n+1},\vect{w}_{1:n}) & \mathcal{K}(w_{n+1},w_{n+1})
    	\end{bmatrix} \right)
    	\end{equation}
    	The GP-UCB process is iterated until the training points (i.e., values of the regularization parameter $ w $), collected by Eq. (\ref{query point}) converge, where the convergence criterion is defined as $ \|w_{n+1}-w_n \|<10^{-2} $.
    	The converged value of $ w $ will be selected as the optimal regularization parameter.

    \subsection{Parallel Computing to Increase Computation Efficiency}

    In the proposed P-DL method, the DNN parameters (i.e., $ \theta_{NN} $) can be estimated by back-propagation algorithms incorporated with optimization methods such as stochastic gradient descent (SGD), adaptive moment estimation (Adam), and other optimization methods. In our work, we engage Adam as the optimizer due to its improved generalization performance \cite{kingma2014adam}. This process involves evaluating the gradients of the loss function at each training point (or collocation point) and updating  $ \theta_{NN} $ iteratively. A large number of collocation points will increase the computational intensity of the training process.  Fortunately, an appealing feature of the P-DL method is that computing tasks are much easier to be operated in parallel than traditional inverse ECG modeling. Here, we propose to scale up the P-DL algorithm by effectively harnessing the computing power of multiple processors collaboratively.
    
      	\begin{figure}[htb]
    	\centering
    	\includegraphics[width=3in]{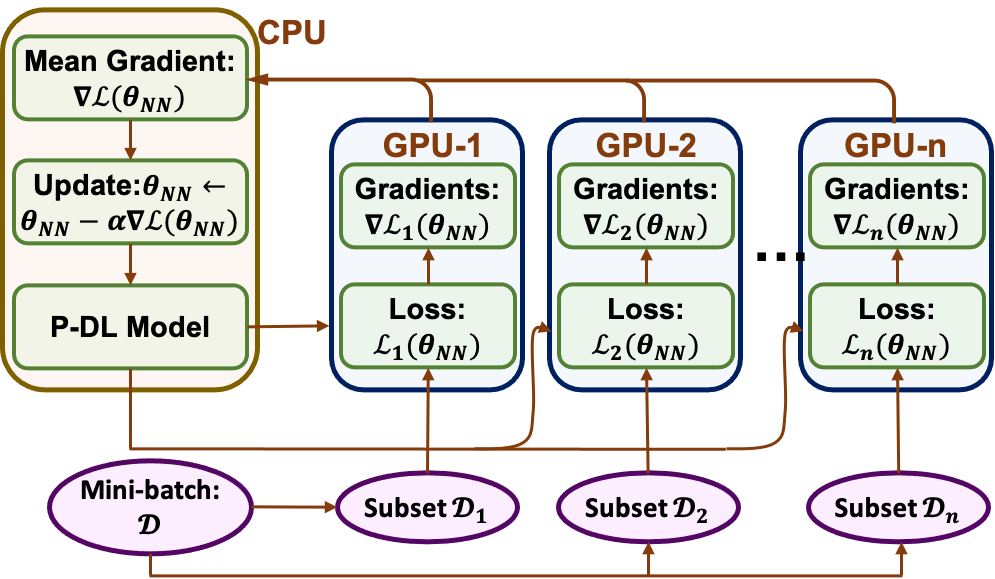}
    	\caption{Scheme for parallel computing.}
    	\label{Fig:parallel}
    \end{figure}
    
    Fig. \ref{Fig:parallel} shows the proposed strategy for parallel computing. To increase the computational parallelism and escape from the local minima \cite{bottou2010large, ge2015escaping}, the whole collocation dataset is divided into mini-batch data (denoted as $ \mathcal{D} $), which are further split for the Adam algorithm into $ n $ subsets during network training, i.e., $ \mathcal{D} = \mathcal{D}_1 \bigcup \mathcal{D}_2 \bigcup \cdots \bigcup \mathcal{D}_{n} $. Thus, the gradient of the loss function with data $ \mathcal{D} $ will be calculated as 
    \begin{eqnarray}
    	\nabla \mathcal{L}(\theta_{NN}) = \frac{1}{| \mathcal{D}|} [ \nabla \mathcal{L}_1(\theta_{NN})+ \cdots + \nabla \mathcal{L}_{n}(\theta_{NN}) ]
    \end{eqnarray}
    where $ \nabla \mathcal{L}_i(\theta_{NN}) = \sum_{j \in \mathcal{D}_i} \nabla \mathcal{L}_i(\theta_{NN}) $, denoting the sum of gradients on subset $ \mathcal{D}_i $. As such, parallel computing can be readily implemented to update network hyperparameters by assigning each subset $ \mathcal{D}_i $ to an individual processor and combining results from many computing units.

       \begin{figure}[H]
      	\centering
      	\includegraphics[width=3.5in]{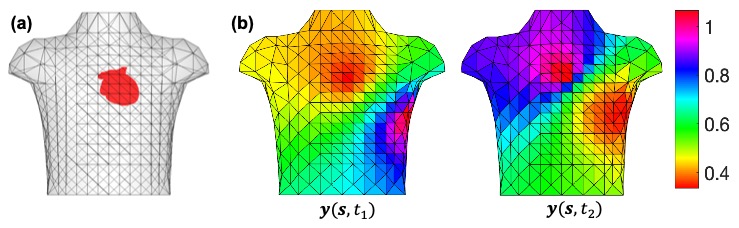}
      	\caption{(a) Illustration of the torso-heart geometry; (b) BSPM at two different time points $t_1$ and $t_2$.}
      	\label{Fig:torso-heart}
      \end{figure}
\section{Experimental Design and Results}
 We validate and evaluate the performance of the proposed P-DL framework in a 3D torso-heart geometry to estimate the HSP from BSPM measurements. As shown in Fig. \ref{Fig:torso-heart}(a), the heart geometry is formed by 1094 nodes and 2184 mesh elements, and the torso surface consists of 352 nodes and 677 mesh elements, which is obtained from the 2007 PhysioNet Computing in Cardiology Challenge \cite{dawoud2008using,goldberger2000physiobank}. This torso-heart geometry will result in a transfer matrix $ \vect{R} $ with a dimensionality of $352\times 1094$. The spatiotemporal HSP and BSPM data are developed from the two-variable cardiac reaction-diffusion system through the procedure detailed in \cite{wang2009physiological}. The reference HSP data $\vect{u}(\vect{s},t)$ consists of $1094\times661$ data points, which are used to evaluate the quality of inverse ECG solution $\vect{\hat{u}}(\vect{s},t)$. In other words, there are 1094 reference time-series data from the heart geometry. Note that the cardinality of the discretized time steps is 661 in the present investigation. The BSPM data $\vect{y}(\vect{s},t)$ has a dimension of $352\times661$, which goes into the data-driven loss (i.e., $\mathcal{L}_{hb}$) to train the P-DL and inversely reconstruct the HSP signal. Fig. \ref{Fig:torso-heart}(b) illustrates the spatiotemporal BSPMs that will be used to predict the HSP.
 
\begin{figure}[H]
   	\centering
   	\includegraphics[width=3.5in]{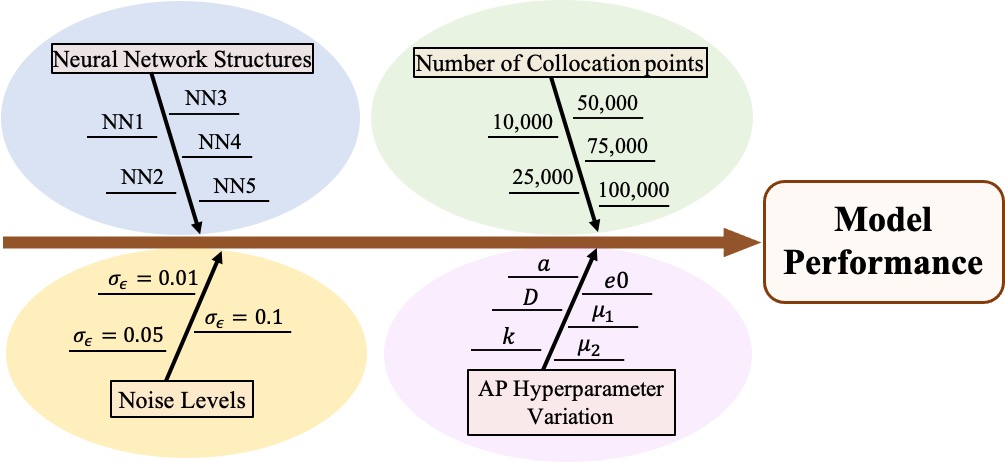}
   	%\captionsetup{justification=centering}
   	\caption{Experimental design for evaluating the performance of the proposed P-DL approach.}
   	\label{Fig:fish}
   \end{figure}

  We investigate the impact of neural network structure, number of collocation points, noise level, and physics model parameters on the prediction performance as shown in Fig. \ref{Fig:fish}. The proposed P-DL approach will be benchmarked with traditional regularization methods including Tikhonov zero-order (Tikh\_0th), Tikhonov first-order (Tikh\_1st), the spatiotemporal regularization (STRE) (see Eq. (\ref{Eq:tikh})-(\ref{Eq:STRE}) for the objective functions), and the physiological-model-constrained Kalman filter (P-KF) method. The model performance will be evaluated according to three metrics: Relative Error ($RE$), Correlation Coefficient ($CC$), and Mean Squared Error ($MSE$), which are defined as
\begin{align}
RE&= \frac{\sqrt{\sum_{\vect{s},t} \| \vect{\hat{u}}(\vect{s},t)-\vect{u}(\vect{s},t) \|^2}}{\sqrt{\sum_{\vect{s},t}\| \vect{u}(\vect{s},t) \|^2}}\\
CC&= \frac{\sum_{\vect{s}}(\vect{\hat{u}}(\vect{s},\cdot)-\vect{\bar{\hat{u}}}(\vect{s},\cdot))^T\cdot (\vect{{u}}(\vect{s},\cdot)-\vect{\bar{{u}}}(\vect{s},\cdot))}{\sum_{\vect{s}}\|(\vect{\hat{u}}(\vect{s},\cdot)-\vect{\bar{\hat{u}}}(\vect{s},\cdot))\|\cdot\|(\vect{{u}}(\vect{s},\cdot)-\vect{\bar{{u}}}(\vect{s},\cdot))\|}\\
MSE&= \frac{1}{N_{st}}\sum_{\vect{s},t} \| \vect{\hat{u}}(\vect{s},t)-\vect{u}(\vect{s},t) \|^2 
\end{align}    
where $N_{st}$ stands for the total number of spatiotemporal instances, $ \vect{u}(\vect{s},t) $ and $ \vect{\hat{u}}(\vect{s},t)$ represent the reference and estimated HSPs, $\vect{\hat{u}}(\vect{s},\cdot)$ and $\vect{{u}}(\vect{s},\cdot)$ denote the estimated and reference time series vectors at spatial location $\vect{s}$, and $\vect{\bar{\hat{u}}}(\vect{s},\cdot)$ and $\vect{\bar{{u}}}(\vect{s},\cdot)$ are the mean of $\vect{\hat{u}}(\vect{s},\cdot)$ and $\vect{{u}}(\vect{s},\cdot)$, respectively.
The metric $RE$ quantifies how large the discrepancy between the estimation $ \vect{\hat{u}}(\vect{s},t)$ and the truth $ \vect{u}(\vect{s},t) $ compared to the magnitude of the true signals. A smaller value of $ RE $ indicates a better model performance. $CC$ characterizes the degree of pattern similarity between $ \vect{u}(\vect{s},t) $ and $ \vect{\hat{u}}(\vect{s},t)$ with $CC=1$ for perfect pattern match and $CC=0$ for zero correlation (i.e., $ \vect{u}(\vect{s},t) $ and $ \vect{\hat{u}}(\vect{s},t)$ are not similar at all). $MSE$ measures the average squared difference between $ \vect{u}(\vect{s},t) $ and $ \vect{\hat{u}}(\vect{s},t)$.
   
  The proposed P-DL algorithm is successfully implemented on TensorFlow, which is a popular platform for deep learning. TensorFlow can be operated on the heterogeneous environment including a multi-core Central Processing Unit (CPU), Graphics Processing Unit (GPU), and Tensor Processing Unit (TPU) \cite{abadi2016tensorflow}. Such cutting-edge hardware developments make parallelization applicable, which significantly increases the computation efficiency in DNN training \cite{shi2016benchmarking}. In our work, we process the P-DL algorithm on the parallel computing platform NVIDIA CUDA, which is a software framework for programs running across different environments such as CPU and GPU. The neural network is carried on TensorFlow-GPU with Python application programming interface (API), and we found that the computation speed is around 6.5 times faster than running on CPU only. Note that the CPU used for the computation is Intel(R) Xeon(R) W-2223CPU @ 3.6GHz. The GPU is NVIDIA Quadro P2200.

    \subsection{Impact of Neural Network Structure on the Prediction Performance}
    \label{Subsec:NN}

    In the present investigation, we study the impact of network structure on the prediction performance of the proposed P-DL method. Specifically, five feedforward fully-connected neural network structures are evaluated, whose structure details are listed in Table \ref{Table:nn}. Each network structure is evaluated by executing P-DL with the same data input and hyperparameters. Note that a Gaussian noise $\epsilon(\vect{s},t)\sim\mathcal{N}(0,\sigma_{\epsilon}^2)$ with standard deviation $\sigma_{\epsilon}=0.01$ is added to BSPM data $ \vect{y}(\vect{s},t) $. GP-UCB algorithm is first implemented to find the best regularization parameter to balance between $\mathcal{L}_{hb}$ and $\mathcal{L}_{ph}$ for each neural network structure. The optimal value $w_{opt}$ for each network structure is shown in Appendix B. 
    
      \begin{table}[H]
     	\caption{Structure details of different neural networks}
     	\label{Table:nn}
     	\vspace{-0.3cm}%Workaround to be conform with the .doc style. Only for table captions.
     	\renewcommand\arraystretch{1.5}
     	\begin{center}
     		\begin{tabular}{cc|ccccc}
     			\hline
     			\multicolumn{2}{l}{}  & NN1 & NN2 & NN3 & NN4 & NN5 \\ \hline
     			
     			\multirow{3}{*} & Layers & 5 & 3 & 5 & 8 & 5 \\ \cline{1-7}
     			
     			& Neurons & 5 & 10 & 10 & 10 & 20 \\ \cline{1-7}
     			
     		\end{tabular}
     	\end{center}
     \end{table}
    
\begin{figure*}
    	\centering
    	\includegraphics[width=7in]{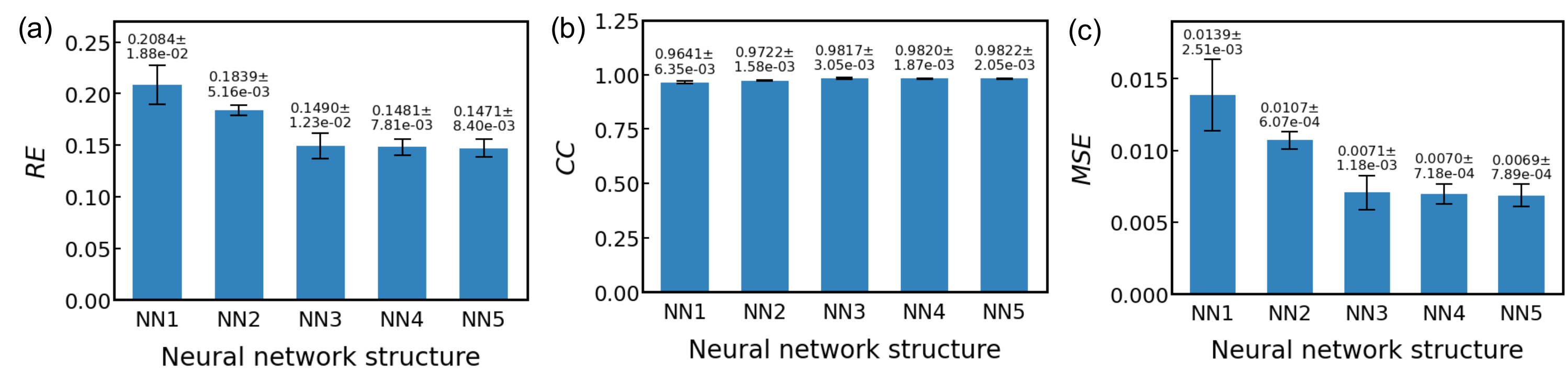}
    	\caption{The variation of (a) $ RE $, (b) $ CC $, {and (c) $MSE$} with respect to neural network structure.}
    	\label{Fig:NN}
    \end{figure*}

    Fig. \ref{Fig:NN}(a-c) compares the $ RE $, $ CC$ and $ MSE$, for different network structures.  Note that the experiments are repeated 10 times, and therefore every obtained $ RE $, $ CC $ {and $MSE$} comes with a corresponding error bar (i.e., one standard deviation of $ RE $, $CC$ {and $MSE$}). The network being too thin (e.g., NN1 with 5 neurons for each layer) or too shallow (e.g., NN2 with only 3 layers) may cause relatively high $ RE $ with a value of ($ 0.2084 \pm 0.0188$) or ($ 0.1839 \pm 0.0052 $), {big $ MSE $ value of ($ 0.0139 \pm 0.0025$) or ($ 0.0107 \pm 0.0006 $),} and low $ CC $ value of ($ 0.9641 \pm 0.0064 $) or ($ 0.9722 \pm 0.0016 $). Thus, a more complex structure is preferred in terms of prediction performance. If we increase the number of neurons in each layer (e.g., NN3 and NN5), or add more layers to the network (e.g., NN4), the resulted $ RE $ and $ MSE $ are significantly reduced. According to Fig.\ref{Fig:NN} (a-c), there is no significant difference among the $RE$, $ CC $, {and $MSE$}'s obtained from P-DL with NN3, NN4, and NN5, indicating further increasing the network complexity may not be necessary. As such, we choose NN3 as the network structure for later experiments.
    
    \begin{figure*}
    	\centering
    	\includegraphics[width=7in]{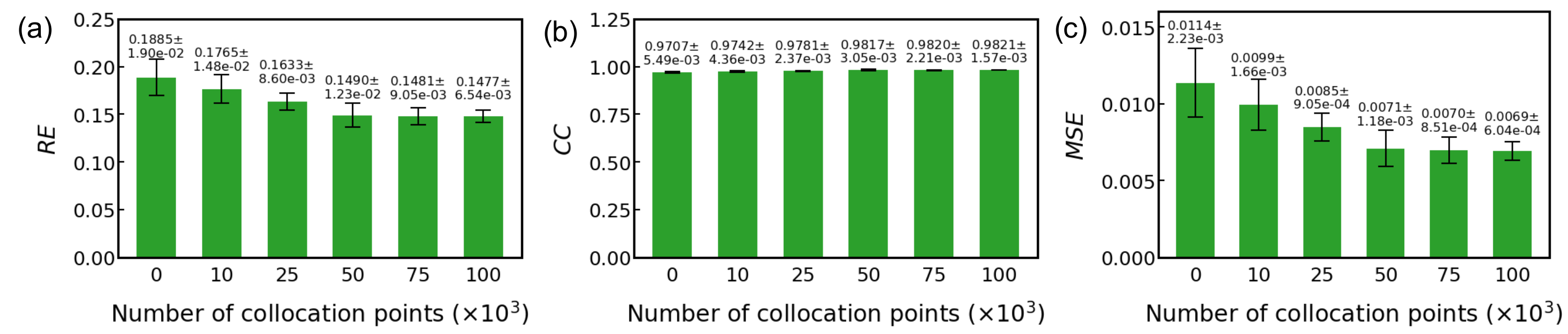}
    	\caption{The variation of (a) $ RE $, (b) $ CC $, {and (c) $MSE$} with respect to the number of collocation points. }
    	\label{Fig:col}
    \end{figure*}

	\subsection{Impact of Collocation Points on the Prediction Performance}
	{To verify the necessity and efficacy of the physics-driven regularization, we further compare the performance of our P-DL model with the data-driven deep learning (DL) method. The loss function of the data-driven DL is the pure data-driven loss defined in Eq. (9). In other words, zero collocation points will be selected to encode the physics-based principle. We also randomly select different numbers of spatiotemporal collocation points to enforce the physics laws imposed by the AP model. Specifically, we choose five numbers of collocation points: $ N_{ph} = 10000, 25000, 50000, 75000, 100000 $ to quantify its impact on the predictive accuracy of the proposed P-DL framework. Similarly, we engage the DNN with 5 layers and 10 neurons in each layer as the model infrastructure. 
		\begin{figure}
			\centering
			\includegraphics[width=2.8in]{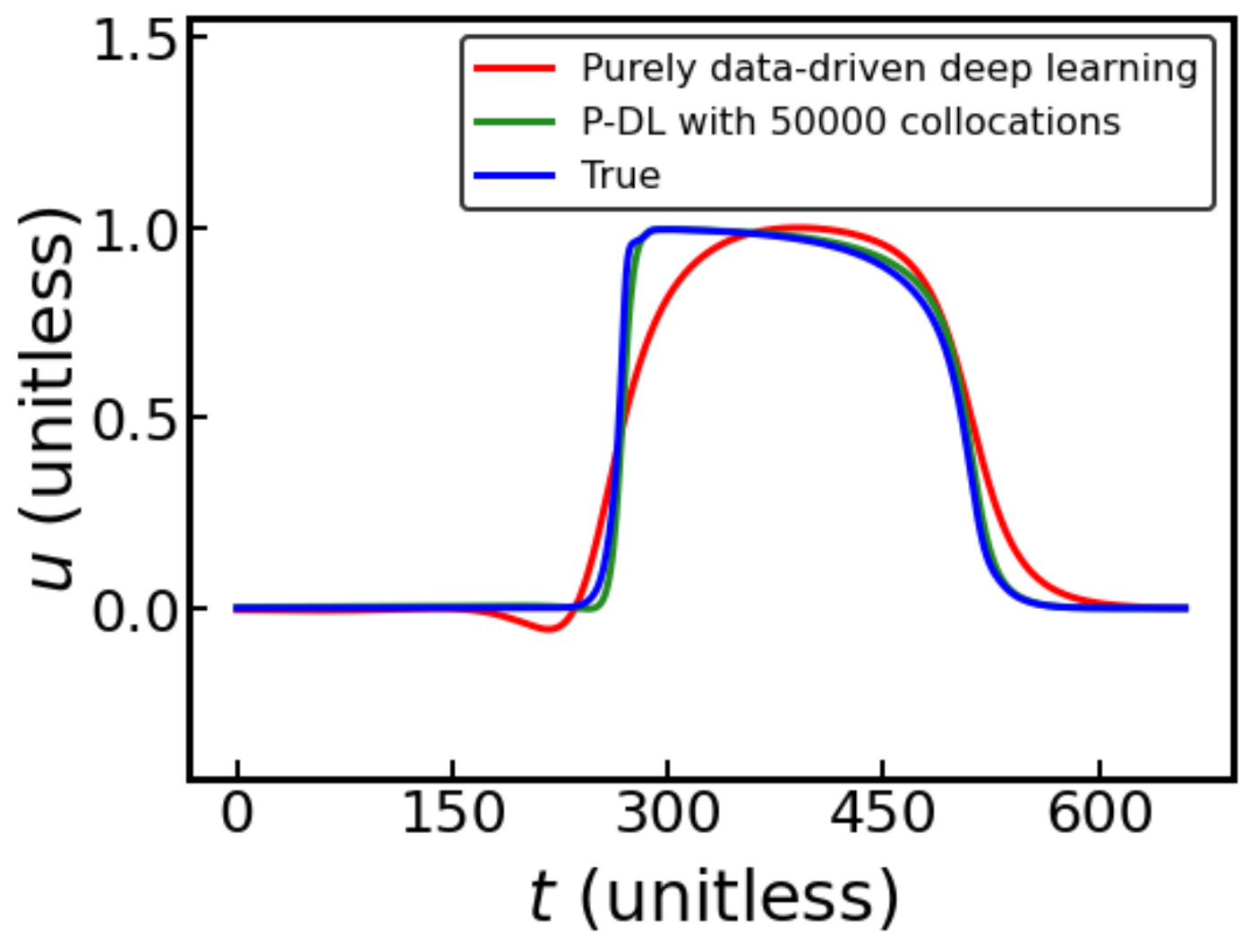}
			\caption{{The estimated HSP evolvement $\vect{\hat{u}}(\vect{s},t)$ from an arbitrary spatial location in the heart generated from pure data-driven DL and P-DL with 50000 collocation points compared with the true evolvement. } }
			\label{Fig:pureevolve}
		\end{figure}

	   \begin{figure*}
	   	\centering
	   	\includegraphics[width=5in]{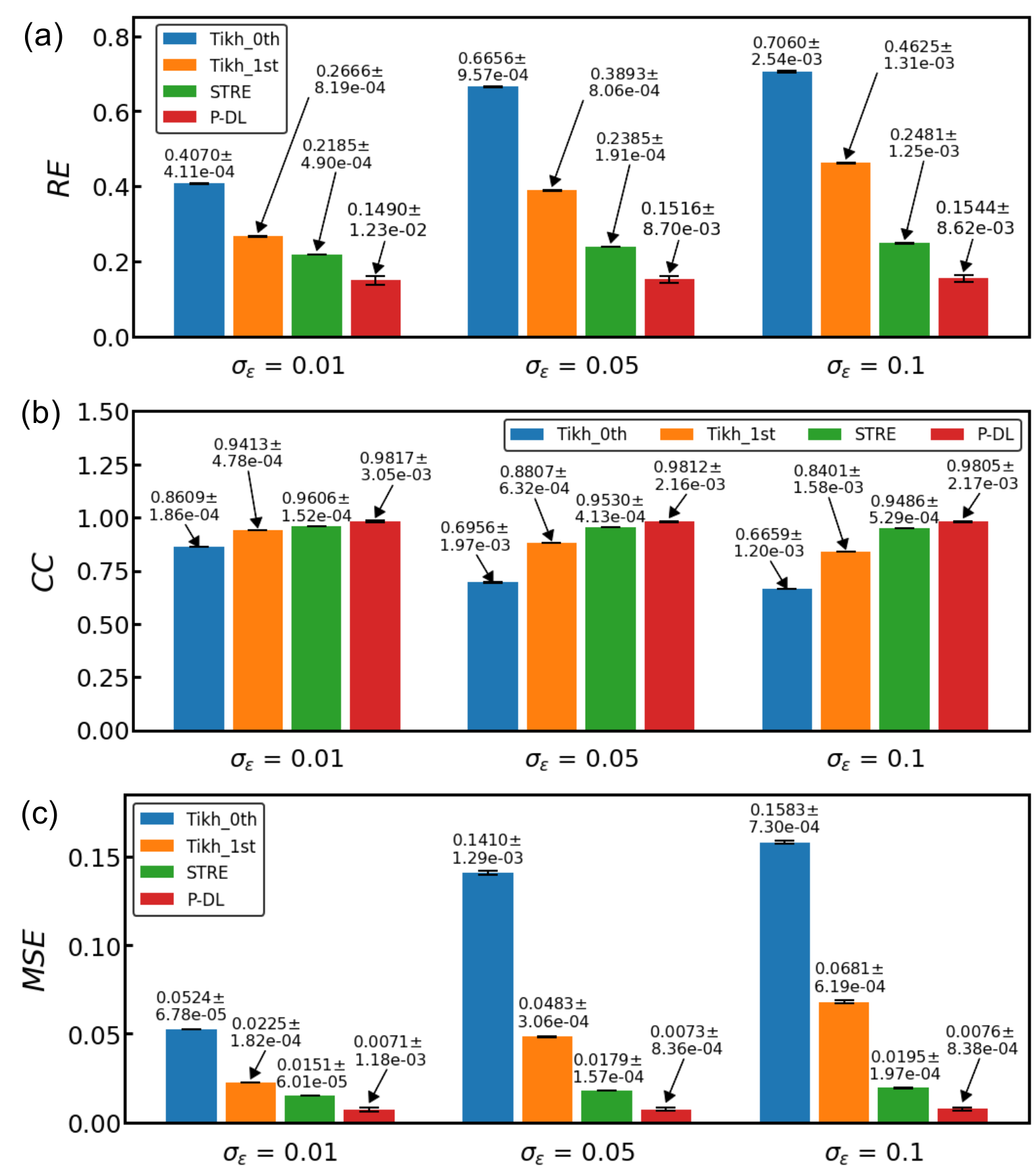}
	   	%	\captionsetup{justification=centering,margin=2cm}
	   	\caption{The comparison of (a) $RE$, (b) $ CC $, {and (c) $MSE$} between the proposed P-DL method and other regularization models (i.e., Tikh\_0th, Tikh\_1st, and STRE) when different noise levels $ \sigma_\epsilon = 0.01,\ 0.05, \  0.1 $ are added to the BSPM $ \vect{y}(\vect{s},t) $.}\label{Fig:comparison}
	   \end{figure*}
	 
	 As shown in Fig. \ref{Fig:col}(a-c), the $RE$ and $MSE$ decrease, and $ CC $ increases as we raise the number of collocation points. Specifically, with $ N_{ph} = 0 $, i.e., the pure data-driven DL has the worst estimation performance, which is due to the fact that the physics principle is absent from the model construction. By increasing $ N_{ph} $ to 50000, $ RE $ and $MSE$ are reduced to ($ 0.1490 \pm 0.0123 $) and ($ 0.0071 \pm 0.0012 $), while $ CC $ increases to ($ 0.9817 \pm 0.0031 $). If we continue increasing the number of collocation points, the estimated $RE$, $MSE$, and $CC$ do not improve significantly, besides it may potentially increase the computational burden. Therefore, we select 50000 collocation points for later numerical experiments. Fig. \ref{Fig:pureevolve} depicts the HSP evolvement at one arbitrary location on the heart geometry yielded from the data-driven DL and the P-DL with 50000 collocation points, compared with true evolvement. The predicted evolvement by the data-driven DL shows a more significant discrepancy from the true evolvement pattern compared with the one estimated by the P-DL model.}

	\begin{figure*}
		\centering
		\includegraphics[width=5in]{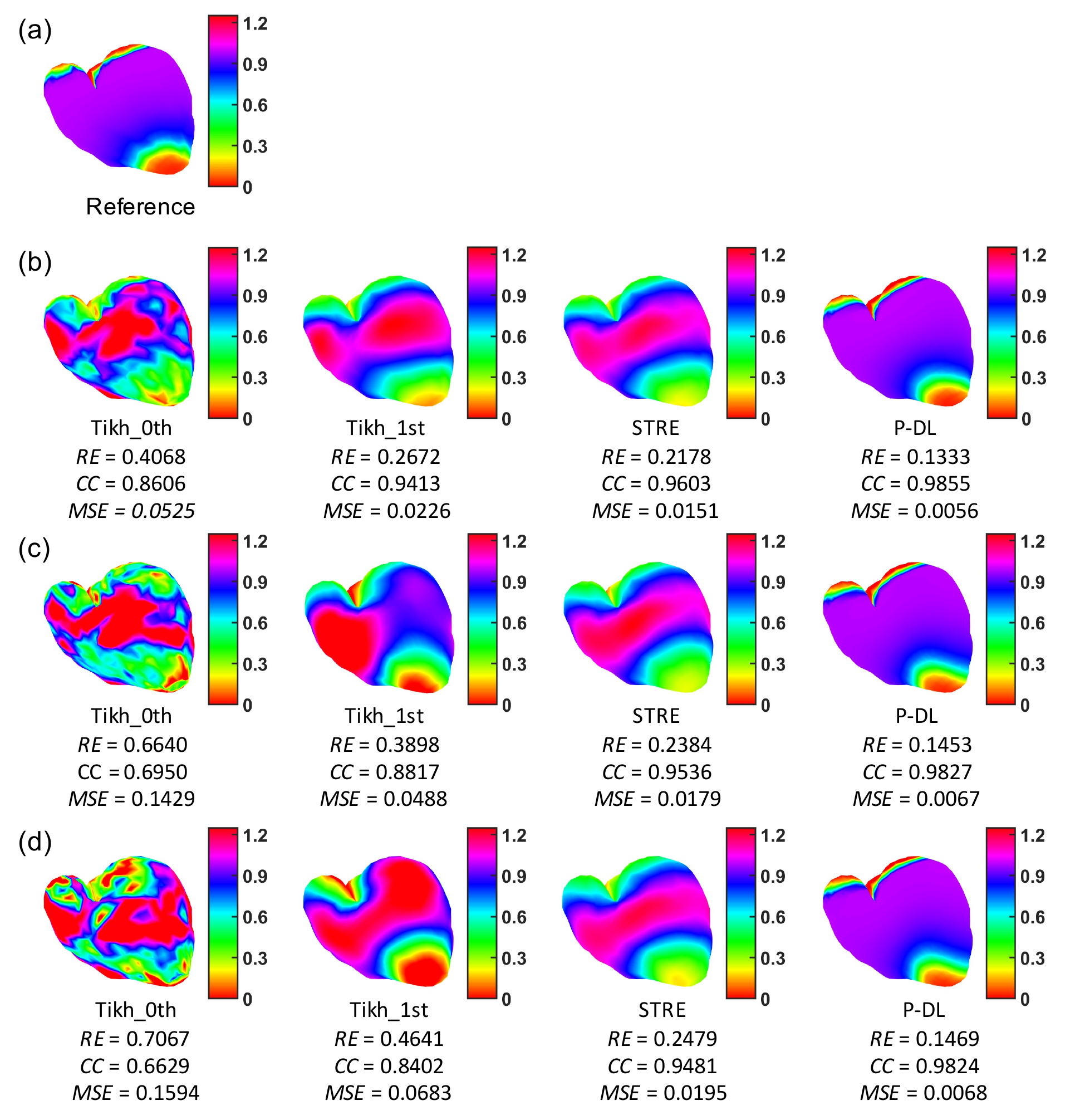}
		\caption{(a) Reference HSP mapping $ \vect{u}(\vect{s},t) $ at $ t=300 $; Estimated mappings $ \vect{u}(\vect{s},t) $ at $ t=300 $ by P-DL and traditional methods (i.e., Tikh\_0th,  Tikh\_1st, and STRE) with noise level of (b) $ \sigma_\epsilon = 0.01 $ (c) $ \sigma_\epsilon = 0.05 $ and (d) $ \sigma_\epsilon = 0.1 $ in BSPM $\vect{y}(\vect{s},t)$. }
		\label{Fig:healthymapping}
	\end{figure*}

   \subsection{Comparison Study with Traditional Regularization Methods}
   We further benchmark the proposed P-DL framework with traditional regularization methods that are commonly used in current practice (i.e., Tikh\_0th, Tikh\_1st, and STRE). Fig. \ref{Fig:comparison} (a-c) shows the comparison of $RE$ , $ CC $, {and $MSE$} obtained from the four methods at different noise levels of $ \sigma_\epsilon = 0.01,\ 0.05,$ and 0.1. As the noise level increases from $ \sigma_\epsilon = 0.01 $ to $ \sigma_\epsilon = 0.1 $, the $ RE $ {and $MSE$} increase monotonically for all the methods. {Specifically, $ RE $ and $MSE$ increase from ($0.4070\pm 0.0004$), ($0.0524\pm 7E-5$) to ($ 0.7060 \pm 0.0025 $), ($0.1583\pm 0.0007$) for Tikh\_0th, 
   from ($ 0.2666\pm 0.0008 $), ($ 0.0225\pm 0.0002 $) to ($ 0.4625 \pm 0.0013 $), ($ 0.0681\pm 0.0006 $) for Tikh\_1st, 
   from ($ 0.2185 \pm 0.0005 $), ($ 0.0151 \pm 0.0002 $) to ($ 0.2481 \pm 0.0013 $), ($ 0.0195 \pm 0.0002 $) for STRE, 
   and from ($ 0.1490 \pm 0.0123 $), ($ 0.0071 \pm 0.0012 $) to ($ 0.1544 \pm 0.0086 $), ($ 0.0076 \pm 0.0008 $) for the P-DL model.} Correspondingly, $ CC $ decreases for each method as the noise level is raised up to $ \sigma_\epsilon = 0.1 $. Overall, the proposed P-DL model generates the smallest $ RE $ {and $MSE$}, and highest $ CC $ compared to other regularization methods regardless of the noise level.

   As illustrated in Fig. \ref{Fig:comparison}, it is also worth noting that the estimated $ RE $ {and $MSE$} obtained from the commonly used Tikh\_0th and Tikh\_1st methods increases dramatically as the bigger noise being added to BSPM data $ \vect{y}(\vect{s},t) $, where the $ RE $ reaches a $ 73.4\% $ and $ 73.5\% $ growth, {and $MSE $ increases by 202.0\% and 202.7\%} as the noise level increases from $ \sigma_\epsilon = 0.01 $ to $ \sigma_\epsilon = 0.1 $, respectively. The STRE model reduces the effect of noise level but still yields an increase level of $ 13.5\% $ in $RE$ {and 29.1\% in $MSE$}. On the other hand, the proposed P-DL method achieves only a $ 3.6\% $ increase in $ RE $ {and $7.0\%$ in $MSE$} as the noise level reaches $ \sigma_\epsilon = 0.1 $. Similarly, $ CC $ drops 26.7\%, 10.7\%, and 1.2\% for Tikh\_0th, Tikh\_1st, and STRE model, respectively, whereas for P-DL model, $ CC $ only declines by 0.12\% as the noise level grows from $ \sigma_\epsilon = 0.01 $ to $ \sigma_\epsilon = 0.1 $, exhibiting extraordinary anti-noise performance. Thus, P-DL model not only accomplishes superior prediction accuracy, but also demonstrates the predictive robustness when encountering a high level of noise in BSPM.

   Fig. \ref{Fig:healthymapping}(a) delineates the true potential mappings on the heart surface. Because the potential distribution is changing over time dynamically, Fig. \ref{Fig:healthymapping} only depicts the mappings at one specific time step, i.e., $ t = 300 $. Note that the HSP $ \vect{u}(\vect{s},t) $ is normalized, and both $u$ and $t$ are unitless. Fig. \ref{Fig:healthymapping}(b-d) show the estimated HSP distribution generated by the four different methods, with noise level of $ \sigma_\epsilon = 0.01,\ 0.05$, and 0.1 added to BSPMs $ \vect{y}(\vect{s},t) $, respectively. The estimated potential mappings by Tikh\_0th, Tikh\_1st, and STRE under each noise level present significantly different color patterns compared with the reference mapping in Fig. \ref{Fig:healthymapping}(a). Typically, under the noise level as large as $ \sigma_\epsilon = 0.1 $, {the proposed P-DL yields the $ RE $ and $MSE$ of 0.1469 and 0.0068, which are remarkably lower than the $ RE $ and $MSE$ estimated by other traditional regularization methods, i.e., 0.7067 and 0.1594 by Tikh\_0th, 0.4641 and 0.0683 by Tikh\_1st, 0.2479 and 0.0195 by STRE, respectively,} leading to a similar HSP pattern with the ground truth shown in Fig. \ref{Fig:healthymapping}(a). Similarly, the P-DL model generates the highest $ CC $ of 0.9824, compared to $ CC $ value of 0.6629, 0.8402, and 0.9481 estimated by Tikh\_0th, Tikh\_1st, and STRE, respectively, indicating P-DL model yields the smallest pattern difference with the real potential distribution. This is due to the fact that the P-DL model successfully incorporates the bio-electrical propagation (i.e., AP model) and boundary condition, yielding a robust inverse ECG solution.

      \begin{figure*}
      	\centering
      	\includegraphics[width=7in]{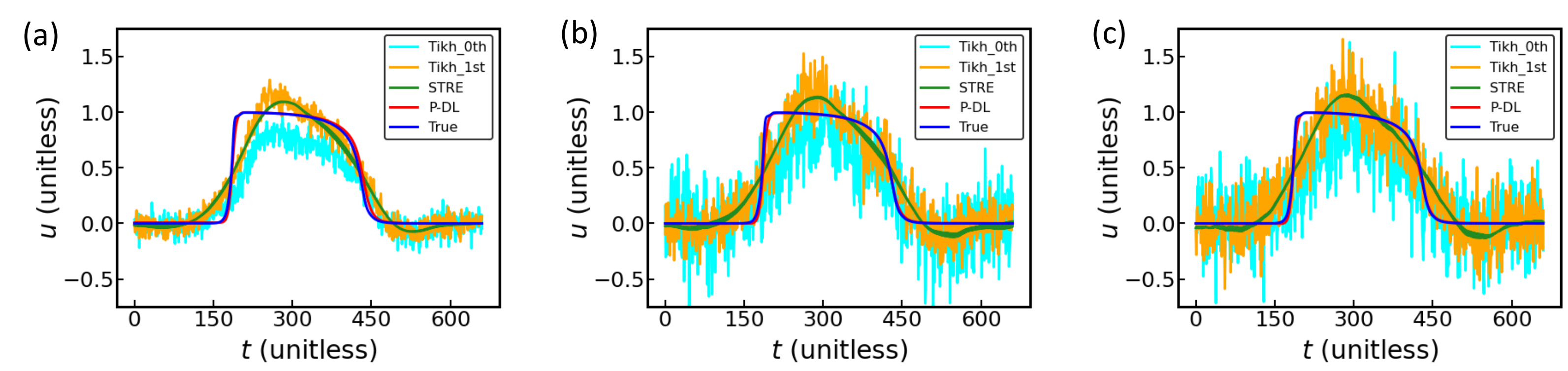}
      	\caption{The comparison of HSP evolvement $ \vect{u}(\vect{s},t) $ over time for one specific spatial location, between the true HSP data and predictions by Tikh\_0th,  Tikh\_1st, STRE, and P-DL under the noise level of (a) $ \sigma_\epsilon = 0.01 $, (b) $ \sigma_\epsilon = 0.05 $, (c) $ \sigma_\epsilon = 0.1 $. }
      	\label{Fig:healthy_oneloc}
      \end{figure*}
  
        \begin{table*}
       %	\color{blue}
       	%\captionsetup{labelfont={color=blue},font={color=blue}}
       	\renewcommand\arraystretch{1.5}
       	\caption{The $ RE $, $ CC $, and $ MSE $ from the P-KF algorithm with different initial conditions. Here,  $ \hat{\vect{u}}_0(\vect{s}) $ and $ \vect{u}_0(\vect{s}) $ denote the specified and true initial condition, respectively; the initialization noise follows a normal distribution, i.e., $\boldsymbol{\epsilon}_{\boldsymbol{u}_{0}}(\boldsymbol{s}) \sim N(0,0.0025)$.}
       	\label{Table: P-KF}
       	\vspace{-0.3cm}%Workaround to be conform with the .doc style. Only for table captions.
       	\begin{center}
       		\begin{tabular}{|m{2.5cm}<{\centering}|m{2cm}<{\centering}|m{2cm}<{\centering}|m{2cm}<{\centering}|m{2cm}<{\centering}|}
       			\hline
       			Initial condition for HSP (i.e., $ \hat{\vect{u}}_0(\vect{s}) $) & $ \vect{u}_0(\vect{s}) $  & $\boldsymbol{u}_{0}(\boldsymbol{s})+\boldsymbol{\epsilon}_{u_{0}}(\boldsymbol{s})$ & Random location initialization & Zero initialization \\
       			\hline
       			$RE$ & $0.1426 \pm 1e-5$ & $ 0.1626\pm 0.0027 $ & $ 0.8940\pm 0.0458 $ & $ 0.9960 \pm 7e-6 $ \\
       			\hline
       			$CC$ & $0.9848 \pm 3e-6 $& $ 0.9803\pm 0.0006 $ & $ 0.3247 \pm 0.0712 $ & $ 0.1563\pm 0.0007 $ \\
       			\hline
       			$ MSE $ & $ 0.0064\pm1e-6 $ & $ 0.0084\pm 0.0003 $ & $ 0.2539 \pm 0.0260 $ & $ 0.3143\pm 0.0028 $ \\
       			\hline
       		\end{tabular}
       	\end{center}
       \end{table*}

   Furthermore, Fig. \ref{Fig:healthy_oneloc} shows the comparison of estimated HSPs $ \vect{\hat{u}}(\vect{s},t) $ evolving over time for one specific spatial location in the heart by the four methods under different noise levels. Note that both Tikh\_0th and Tikh\_1st generate noisy estimation whose noisy pattern becomes more significant when  $ \sigma_\epsilon $ increases from 0.01 to 0.1. The STRE method smoothens out the noise over time in the estimated signal due to the imposed extra temporal regularization, but the temporal regularization does not reflect the physics-based prior knowledge of cardiac electrodynamics, leading to a morphology deviation from the true HSP signal. In contrast, by incorporating the physics-based principles, our proposed P-DL model yields a smooth and robust estimation that matches the morphologic pattern of the true HSP signal well under all the three noise levels as illustrated in Fig. \ref{Fig:healthy_oneloc}(a-c). 
   
   {\subsection{Comparison Study with the Physics-constrained Statistical Method}
    Wang \textit{et al.} proposed to solve the inverse ECG model by fusing the physics-based cardiac model (i.e., the AP model) into the Kalman filter (KF) algorithm (i.e., P-KF) \cite{wang2009physiological}. 
    The detailed algorithm is shown in Appendix C. Note that the prediction accuracy of the P-KF algorithm depends heavily on the assumptions about the initial conditions (i.e., the initial distribution of heart electrical potential $ \vect{u}_0(\vect{s}) $), which is a common problem in KF-based methods \cite{zhao2020trial}. However, the initial potential distribution, i.e., $ \vect{u}_0(\vect{s}) $, at the beginning of the cardiac cycle, depends on the anatomical location of the earliest excitation in the heart, which is difficult to accurately obtain in clinical practice. An inaccurately specified initial condition will significantly reduce the prediction accuracy of the inverse ECG solution. Table \ref{Table: P-KF} summarizes the performance of P-KF under different initial conditions with a noise level of 0.01 added to the BSPM. Note that the hyperparameters in this algorithm are adopted from \cite{wang2009physiological}. The experiments are repeated 10 times for each initial condition.
    
    According to Table \ref{Table: P-KF}, when the initial potential mapping is assigned with the true condition, i.e., $ \widehat{\vect{u}}_0(\vect{s}) = \vect{u}_0(\vect{s}) $, the P-KF algorithm generates the smallest $ RE $ ($ 0.1426\pm1e-5 $) and $ MSE $ ($ 0.0064\pm1e-6 $), and the highest $ CC $ ($ 0.9848\pm3e-6 $), which is comparable with our P-DL method with $ RE $, $ MSE $, and $ CC $ of ($ 0.1490\pm0.0123 $), ($ 0.0071\pm0.0012 $), and ($ 0.9817\pm0.0031 $), respectively. It is also worth mentioning that statistical test shows that there is no significant difference of the $RE$ generated from the P-KF and P-DL (see Appendix D for more detail). However, if the P-KF algorithm is initialized with an inaccurate initial condition, the estimation accuracy will be significantly reduced. For example, if an initialization noise level of $\sigma_{\boldsymbol{\epsilon}_{\vect{u}_{0}}}=0.05 $, i.e., $ \boldsymbol{\epsilon}_{\vect{u}_{0}}(\vect{s}) \sim N(0,0.0025) $, is added to the overall potential mapping at $ t=0 $ (i.e., 
    $\widehat{\boldsymbol{u}}_{0}(\boldsymbol{s})=\boldsymbol{u}_{0}(\boldsymbol{s})+\boldsymbol{\epsilon}_{\boldsymbol{u}_{0}}(\boldsymbol{s})$), the $ RE $ and $ MSE $ rise to ($ 0.1626\pm0.0027 $) and ($ 0.0084\pm0.0003 $), and $ CC $ drops to ($ 0.9803\pm0.0006 $). A random initialization (i.e., randomly picking the earliest excitation on the heart geometry)  can drastically increase the $ RE $ to ($ 0.8940\pm0.0458 $) and $ MSE $ to ($ 0.2539\pm0.0260 $), and reduce the $ CC $ to ($ 0.3247\pm0.0712 $). Similarly, under the zero initialization (i.e., $ \widehat{\boldsymbol{u}}_{0}(\boldsymbol{s})=0 $), the resulted $ RE $ and $ MSE $ are as high as ($ 0.9960\pm7e-006 $) and ($ 0.3143\pm0.0028 $) respectively, and $ CC $ decreases to ($ 0.1563\pm0.0007 $). 
    
     \begin{table*}
    	%\color{blue}
    	%\captionsetup{labelfont={color=blue},font={color=blue}}
    	\renewcommand\arraystretch{1.5}
    	\caption{The computational time for different methods.}
    	\label{Table: efficiency}
    	\vspace{-0.3cm}%Workaround to be conform with the .doc style. Only for table captions.
    	\begin{center}
    		\begin{tabular}{|p{2cm}<{\centering}|p{1.5cm}<{\centering}|p{1.5cm}<{\centering}|p{1.5cm}<{\centering}|p{1.5cm}<{\centering}|p{1.5cm}<{\centering}|}
    			\hline
    			Method & Tikh\_0th & Tikh\_1st & STRE & P-KF & P-DL \\
    			\hline
    			Running Time & 0.31 s & 0.46 s & 2 h 26 min & 3 h 10 min & 19 min\\
    			\hline
    		\end{tabular}
    	\end{center}
    \end{table*}

      \begin{table*}
     	\renewcommand\arraystretch{1.5}
     	\caption{The influence of variations in AP model parameters ($ \pm 10 \% $) on the $ RE $, $ CC $, {and $MSE$} provided by the P-DL.}
     	\label{Table:hyperparameter}
     	\vspace{-0.3cm}%Workaround to be conform with the .doc style. Only for table captions.
     	\begin{center}
     		\begin{tabular}{p{1.0cm}<{\centering}|p{1.3cm}<{\centering}|p{1.3cm}<{\centering}p{1.3cm}<{\centering}p{1.3cm}<{\centering}p{1.3cm}<{\centering}p{1.3cm}<{\centering}p{1.3cm}<{\centering}}
     			\hline
     			\multicolumn{2}{l}{}  & $ a $ & $ k_r $ & $ D $ & $ e0 $ & $ \mu_1 $ & $ \mu_2 $ \\ \hline
     			
     			\multirow{2}*{$ RE $} & $ -10 \% $ & 0.1681 & 0.1525 & 0.1665 & 0.1620 & 0.1625 & 0.1777 \\ \cline{2-8}
     			
     			& $ +10 \% $ & 0.1550 & 0.1797 & 0.1647 & 0.1615 & 0.1949 & 0.1701 \\ \hline
     			
     			\multirow{2}*{$ CC $} & $ -10 \% $ & 0.9768 & 0.9809 & 0.9773 & 0.9785 & 0.9783 & 0.9742 \\ \cline{2-8}
     			
     			& $ +10 \% $ & 0.9804 & 0.9738 & 0.9778 & 0.9786 & 0.9690 & 0.9763 \\ \cline{1-8}
     			
     			\multirow{2}*{$ MSE $} & $ -10 \% $ & 0.0090 & 0.0074 & 0.0088 & 0.0083 & 0.0084 & 0.0100 \\ \cline{2-8}
     			
     			& $ +10 \% $ & 0.0076 & 0.0102 & 0.0086 & 0.0082 & 0.0120 & 0.0092 \\ \cline{1-8}
     			
     		\end{tabular}
     	\end{center}
     \end{table*}
    As such, the performance of P-KF is comparable to our P-DL model only when the true initialization condition is provided (i.e., $ \vect{\hat{u}}_0(\vect{s}) $=$ \vect{u}_0(\vect{s}) $). Incorrect or inaccurate initial conditions will significantly deteriorate the model performance. However, in clinical reality, the knowledge of the true initial condition of HSP mapping is difficult to track accurately, which makes P-KF less applicable in clinical practice. In contrast, our P-DL model does not require any prior information of initial HSP mapping at $ t=0 $. 
    
    Moreover, the KF-based model is more computational-intensive because the algorithm is iterated along the discretized time domain and involves repeated numerical evaluations of the AP model and large matrix manipulation at each time instance. Specifically, for each time instance, the discretized AP model is evaluated $2N+1$ times to generate the \textit{priori} prediction of HSP, where $N$ is the total number of the discretized nodes on the 3D heart geometry (see more details at Appendix C). Table \ref{Table: efficiency} further summarizes the program running time for each method.  It takes about 19 minutes for the proposed P-DL to converge and generate the inverse ECG solution, which is much more computation-efficient compared with the P-KF method (i.e., 3 hours and 10 minutes) and the STRE method (i.e., 2 hours and 26 minutes). Note that Tikh\_0th and Tikh\_1st can generate the inverse results within a second. This is due to the fact that both Tikh-based methods have closed-form solutions and only focus on regularizing the spatial structure of the inverse ECG solution but ignore the patterns in the temporal domain. 
    
}

    \subsection{Impact of Electrophysiological Model Parameters on the Prediction Performance}
   In the case that AP model could not precisely describe the underlying cardiac electrodynamics, it is necessary to investigate how the variations in the model parameters will influence the predictive accuracy of P-DL model. Table \ref{Table:hyperparameter} lists the $ RE $ and $ CC $ when the individual hyperparameter (i.e., $ a $, $ k_r $, $ D $, $ e0 $, $ \mu_1 $, $ \mu_2 $) deviates from the reference value by $ \pm 10\% $, given that a noise level of $ \sigma_\epsilon = 0.01 $ is added to BSPM $ \vect{y}(\vect{s},t) $. Note that $ RE $, $MSE$, and $ CC $ evaluated with the reference parameter setting (i.e., $ a=0.1 $, $ D=10 $, $ k_r=8 $, $ e0=0.002 $, $ \mu_1=\mu_2=0.3 $) are ($ 0.1490 \pm 0.0123 $), ($ 0.0071 \pm 0.0012 $), and ($ 0.9817 \pm 0.0031 $) as shown in Fig. \ref{Fig:comparison}. 
   
   After designated  $ \pm 10\% $ variations to the AP parameters, {the overall prediction performance deteriorates with an increase in $ RE $ and $MSE$ and a decrease in $ CC $. Nevertheless, as shown in Table \ref{Table:hyperparameter}, $ RE $ and $MSE$ predicted by the proposed P-DL model with $ \pm 10\% $ variations in the AP hyperparameters still rank the smallest compared with traditional regularization methods, i.e., ($ 0.4070 \pm 0.0004 $) and ($ 0.0524 \pm 7e-5 $) by Tikh\_0th, ($ 0.2666 \pm 0.0008 $) and ($ 0.0225 \pm 0.0002 $) by Tikh\_1st, and ($ 0.2185 \pm 0.0005 $) and ($ 0.0151 \pm 6e-5 $) by STRE, respectively.} Similarly, the $ CC $ estimated by the P-DL model still scores the highest value among the four methods, i.e., ($ 0.8609 \pm 0.0002 $) by Tikh\_0th, ($ 0.9413 \pm 0.0005$) by Tikh\_1st, and ($0.9606 \pm 0.0002 $) by STRE even when the model is provided with an inaccurate AP parameters. Although the AP model with imperfect parameter setting is encoded as the physics constraint into the P-DL model, the body-heart transformation loss $ \mathcal{L}_{hb} $ also plays an important role in optimizing the neural network. These two parts interact with each other, making the P-DL model robust against the measurement noise and parameter uncertainties.
   
{
\section{Discussion}   
     \subsection{Advantages of the Proposed P-DL Method in Inverse ECG Modeling }
         The advantages of the P-DL method in solving the inverse ECG problem over the statistical models are listed as follows:
         
         (1) \textbf{The physics prior knowledge can be easily incorporated into DNN}: The spatiotemporal cardiac electrodynamics is generally described by the physics-based principles in terms of partial differential equations (PDEs). Due to the rapid advancements in deep learning, the physics-based PDEs (i.e., the physics-based cardiac model) can be easily integrated with data-driven loss in the DNN through automatic differentiation. The physics-based cardiac model in the P-DL will serve as a spatiotemporal regularization to respect the evolving dynamics of cardiac electrical signals. On the other hand, it is difficult to encode the physics law into traditional statistical modeling. In fact, in order to achieve high-quality prediction with statistical methods, researchers have to regularize the inverse ECG solution by delicately designing a proper constraint, usually by mathematically developing a spatial operator to characterize the correlation in the space domain and/or adding temporal regularization to increase the prediction robustness in the time domain. In other words, traditional regularization methods in inverse ECG modeling focus on exploiting the data structure (i.e., data-driven regularization), but ignore existing physics prior knowledge for regularization. As such, the proposed P-DL-based inverse ECG method achieves a better prediction performance compared with traditional regularization methods as shown by the experimental results in Fig. \ref{Fig:comparison}-\ref{Fig:healthy_oneloc}.
         
         (2) \textbf{Computation Efficiency}: Another key attractive feature of deep learning-based modeling is that computing tasks are much easier to be executed in parallel, which will increase the computation efficiency by harnessing the computing power of multiple processors collaboratively. However, solving the statistical models with spatiotemporal regularization (e.g., the STRE or P-KF method) requires either large matrix manipulation, which can be computationally prohibitive for large-scale inverse ECG modeling, or the development of complex optimization algorithms, which can be exceedingly mathematically heavy, as shown in Table \ref{Table: efficiency}.

     \subsection{Training the P-DL Model}  
       The proposed P-DL network is trained by solving the physics-encoded loss function in Eq. (\ref{Eq: PDL}). Specifically, the training data comes from the body surface measurements (i.e., BSPM $y(\vect{s},t)$), which is related to the HSP through a transfer matrix $\vect{R}$, and the spatiotemporal collocation points to encode the physics-based principles (i.e., the AP model). Note that we did not use the traditional training-validation-testing split to train and evaluate the P-DL model. This is due to the fact that the DL implementation of the inverse ECG modeling is different from traditional DL predictive models. The ultimate goal of inverse ECG modeling is to estimate the HSP from BSPM. The data directly from the object of interest (i.e., the heart surface data $\vect{u}(\vect{s},t)$) does not go into the training set or serve as input of the network. As such, it will not introduce the common problem of overfitting as in traditional machine learning. Additionally, because the electrical signals from each heart node affect the signals on the entire body surface, the electrical potentials either on the heart or body surface cannot be analyzed locally. In other words, the HSP cannot be reliably estimated by a local portion of the BSPM. As such, training-validation-testing split is not generally applicable in inverse ECG modeling \cite{wang2009physiological, messnarz2004new, ghosh2009application,shou2010epicardial}. The quality of inverse solution is commonly evaluated by comparing the estimations $\hat{\vect{u}}(\vect{s},t)$ with the reference signals $\vect{u}(\vect{s},t)$ on the heart surface as shown in existing literature \cite{wang2009physiological, messnarz2004new, ghosh2009application,shou2010epicardial}.
        
      \subsection{Define New Network Structures with Physics-based Models}
      \label{Discussion}
      In the present investigation, we engage a fully-connected feed-forward neural network (NN) structure to solve the inverse ECG problem by defining a unique loss function that fuses the data-driven transformation loss and physics-based loss. In the existing literature, several approaches have been developed to leverage the physics-based knowledge to construct more complicated NN architectures \cite{willard2020integrating}. For example, the NN structures can be adapted by defining intermediate physics variables to produce physics-consistent predictions. Daw \textit{et al.} \cite{daw2020physics} designed a special long-short-term-memory (LSTM) NN by introducing a monotonicity-preserving structure into the LSTM cell, which can help to predict physically consistent densities in lake temperature modeling. Additionally, invariances and symmetries of the systems can be leveraged to design a physics-based NN structure. For instance, Wang \textit{et al.} \cite{wang2020incorporating} incorporated the system symmetries into the convolutional NN (CNN) to improve the accuracy and generalizability of DNN-based predictive models for learning physical dynamics. The NN architecture can also be modified by encoding physics-based domain-specific knowledge. For example, Sadoughi \textit{et al.} \cite{sadoughi2019physics} designed a CNN by incorporating domain-specific knowledge of bearings and their fault characteristics for the defect detection of rolling element bearings.
      
      Designing new NNs with the physics-based cardiac model can be challenging for inverse ECG modeling. It is difficult to find a meaningful intermediate physics variable that is able to improve the overall predictive performance. Additionally, the AP model that describes the cardiac electrodynamics is governed by non-linear PDEs over the complexly structured heart geometry, which does not involve simple invariance or symmetric properties to be used to design effective structures. Thus, in this paper, we focus on incorporating the physics prior knowledge (i.e., the AP model) into the DNN by defining a loss function that combines both the data-driven and physics-based loss. Specifically, the AP model serves as a physics-based spatiotemporal regularization to respect the spatiotemporal evolving dynamics of cardiac signals to improve the predictive performance in inverse ECG modeling. Experimental results demonstrate the superior performance of the proposed P-DL framework compared with existing approaches in solving the inverse ECG problem. In our future work, we will focus on leveraging the physics-based cardiac model to design new network structures to further improve the predictive performance of the P-DL for inverse ECG modeling.
      
}
       
\section{Conclusions}
    
   In this paper, we propose a physics-constrained deep learning (P-DL) strategy to predict the heart-surface electrical signals from body-surface sensor measurements. First, we encode the underlying physics law of cardiac electrodynamics into a DNN to solve the inverse ECG model. Second, in order to balance between the body-heart transformation (i.e., data-driven loss) and physics-based loss, we defined a balance metric as an indicator to efficiently evaluate the predictive performance given different regularization parameters. Third, we propose to solve for the optimal regularization parameter using the GP-UCB algorithm. Finally, we propose to implement the P-DL with TensorFlow-GPU to increase the computation efficiency. We validate and evaluate the performance of the P-DL framework in a 3D torso-heart system to estimate HSP from BSPM data. Numerical experiments demonstrate that the proposed P-DL model is robust against measurement noise and other uncertainty factors. The inverse ECG solution by the P-DL method significantly outperforms existing regularization methods, i.e., Tikhonov zero-order, Tikhonov first-order, STRE model, and physiological-model-constrained Kalman filter. The presented P-DL framework can be broadly implemented to investigate other spatiotemporal high-dimensional systems such as thermodynamics and tumor-growth systems.

{ \section*{Appendix}
\label{appendix}
\subsection{Automatic Differentiation}
\label{automatic_differenciation}
   The partial differential terms in the AP model can be easily calculated by automatic differentiation with no approximation error \cite{baydin2018automatic, raissi2019physics}, which is widely used in DL to compute the derivatives by applying the chain rule repeatedly. Automatic differentiation consists of two major steps: first compute the values of all variables through a forward pass and second calculate the derivatives through a backward pass. We use a simple network example in Fig. \ref{Fig:AD} to demonstrate this technique. The network consists of one hidden layer with inputs $ x $ and $ t $, and one output $ u $. Table \ref{Table:AD} shows the forward and backward pass of automatic differentiation to compute the partial derivatives of $ \frac{\partial u}{\partial x } $ and $ \frac{\partial u}{\partial t} $ at the point $(x,t)=(2,3)$.
      \begin{figure}
      	\centering
      	\includegraphics[width=2.2in]{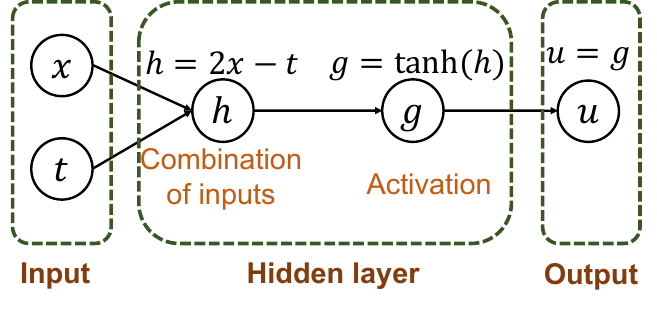}
      %	\captionsetup{labelfont={color=blue},font={color=blue}}
      	\caption{A simple network example to illustrate automatic differentiation.}
      	\label{Fig:AD}
      \end{figure}

\subsection{The optimal regularization parameter $w_{opt}$}
\label{gpucb}
   Table \ref{Table:w_opt} lists the optimal parameter value $w_{opt}$ and the iteration number to converge for each NN structure. The GP-UCB algorithm takes 20 iterations to converge for the structure NN1, NN2, and NN3, yielding the optimal parameter value of 0.23, 0.29, and 0.44, respectively. The GP-UCB for the P-DL with the structure of NN4 and NN5 takes 17 and 19 iterations to converge, with the value of 0.56 and 0.33 for $w_{opt}$, respectively. Fig. \ref{Fig:gpconverge} shows an example of the convergence of GP-UCB. The regularization parameter $ w $ converges after 20 iterations for the structure of NN2.

   \begin{table}
   	%\color{blue}
   	%\captionsetup{labelfont={color=blue},font={color=blue}}
   	\renewcommand\arraystretch{2.8}
   	\caption{The computation of the forward and backward pass of the automatic differentiation example as shown in Fig. \ref{Fig:AD}.}
   	\label{Table:AD}
   	%\vspace{-0.3cm}%Workaround to be conform with the .doc style. Only for table captions.
   	\begin{center}
   	\begin{tabular}{|p{3cm}|p{5cm}|}
   		\hline
   		Forward pass & Backward pass  \\
   		\hline
   		$x=2, t=3$ & $ \dfrac{\partial u}{\partial u }=1 $ \\
   		\hline
		$ h = 2x-t=1$  & $ \dfrac{\partial u}{\partial g }=1 $ \\
		$ g = \tanh(h) =0.7616 $ & $ \dfrac{\partial u}{\partial h}=\dfrac{\partial u}{\partial g} \dfrac{\partial g}{\partial h} = \dfrac{\partial u}{\partial g} \text{sech}^2(h) = 0.4184 $ \\ 
   		\hline
   		$ u = g=0.7616 $  & $ \dfrac{\partial u}{\partial x}=\dfrac{\partial u}{\partial h} \dfrac{\partial h}{\partial x} = 0.4184 \times 2 = 0.8368 $ \\
   		& $ \dfrac{\partial u}{\partial t}=\dfrac{\partial u}{\partial h} \dfrac{\partial h}{\partial t} = 0.4184 \times (-1) = -0.4184 $\\
   		\hline
   	\end{tabular}
   \end{center}
   \end{table}

   \begin{table}
   %	\color{blue}
   	%\captionsetup{labelfont={color=blue},font={color=blue}}
   	\renewcommand\arraystretch{1.5}
   	\caption{The optimal regularization parameter $w_{opt}$ and the iteration number in GP-UCB for each NN structure.}
   	\label{Table:w_opt}
   	%\vspace{-0.3cm}%Workaround to be conform with the .doc style. Only for table captions.
   	\begin{tabular}{p{1.5cm}<{\centering}|p{1.0cm}<{\centering}|p{1.0cm}<{\centering}|p{1.0cm}<{\centering}|p{1.0cm}<{\centering}|p{1.0cm}<{\centering}}
   		\hline
   		NN structure & NN1 & NN2 & NN3 & NN4 & NN5 \\
   		\hline
   		$w_{opt}$ & 0.23 & 0.29 & 0.44 & 0.56 & 0.33\\
   		\hline
   		Iterations & 20 & 20 & 20 & 17 & 19\\
   		\hline
   	\end{tabular}
   \end{table}

\begin{figure}[H]
	\centering
	\includegraphics[width=2.0in]{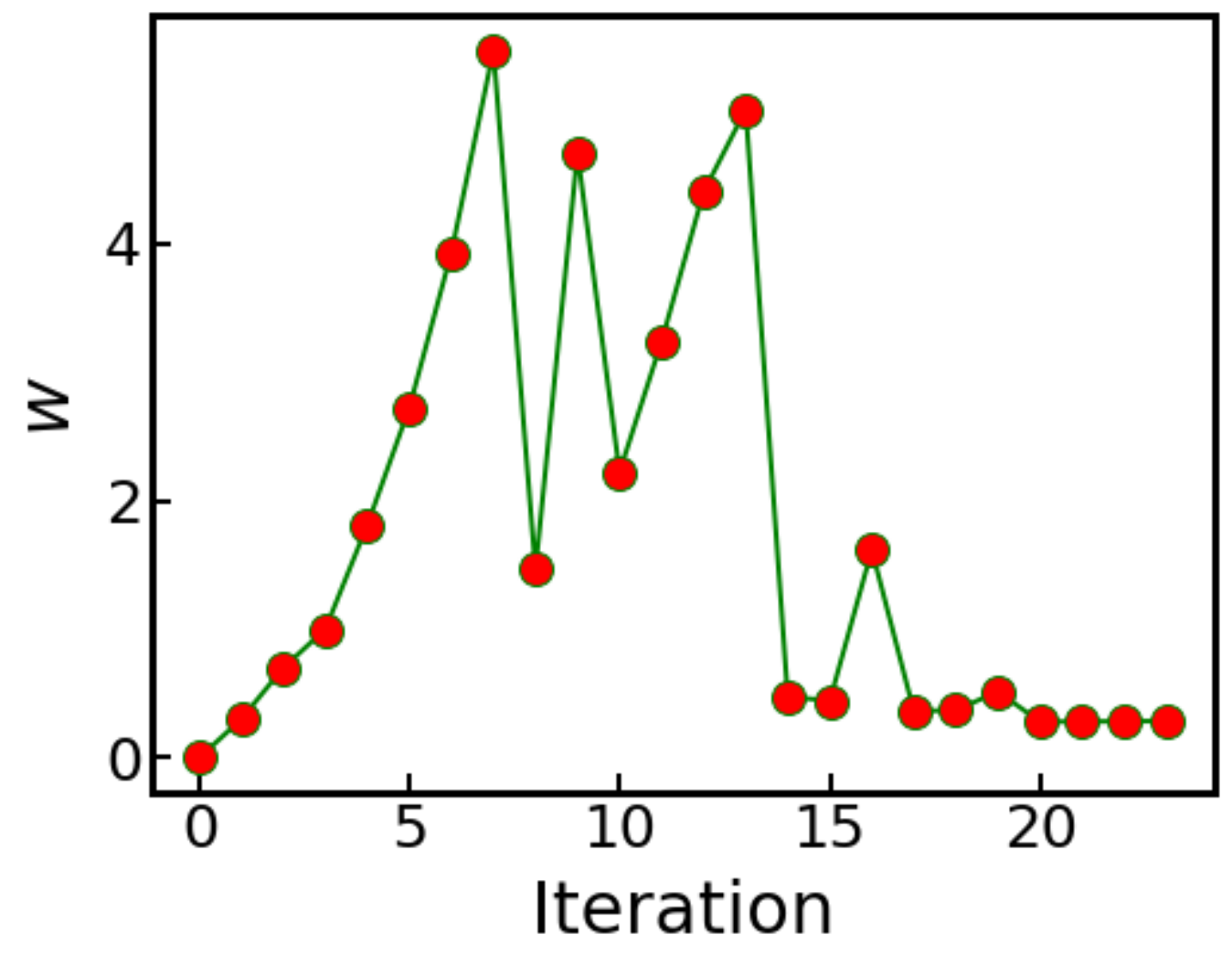}
%	\captionsetup{labelfont={color=blue},font={color=blue}}
	\caption{The illustration of the convergence of GP-UCB.}
	\label{Fig:gpconverge}
\end{figure}

\subsection{Physiological-Model-constrained Kalman filter (P-KF)}
\label{kfalgorithm}
Wang \textit{et al.} \cite{wang2009physiological} proposed a P-KF statistical framework to estimate the cardiac electrical potential from BSPM. The state space representation of the inverse ECG model in the P-KF framework is given as:
\begin{equation}
	\begin{aligned}
		\left(\boldsymbol{u}_{t}, \boldsymbol{v}_{t}\right) &=\mathcal{F}\left(\boldsymbol{u}_{t-1}, \boldsymbol{v}_{t-1}\right)+\boldsymbol{\phi}_{t} \\
		\boldsymbol{y}_{t} &=\boldsymbol{R} \boldsymbol{u}_{t}+\boldsymbol{\varphi}_{t}
	\end{aligned}
\end{equation}
where $ \mathcal{F} $ denotes the AP model, and $\boldsymbol{\phi}_{t} \sim \mathcal{N}\left(\mathbf{0}, \boldsymbol{Q}_{\phi}\right)$ and $\boldsymbol{\varphi}_{t} \sim \mathcal{N}\left(\mathbf{0}, \boldsymbol{M}_{\varphi}\right)$ stand for the model uncertainty and data error at time $ t $, respectively. In order to solve the inverse ECG problem, this method sequentially estimated the electric potential of the heart (i.e., $ \vect{u}_t $) by iteratively assimilating the body-surface data (i.e., $ \vect{y}_t $) into the solution procedure.

Specifically, at each iteration, the P-KF first tackles the nonlinearity of the physics-based cardiac model using the unscented transformation (UT) that is adapted from the unscented Kalman filter (UKF). A set of sigma points $\left\{\vect{u}_{t-1, i}\right\}_{0}^{2 N}$ (where $N$ is the dimension of the state vector $\vect{u}_{t-1}$, i.e., the total number of the discretized spatial nodes on the heart geometry) is generated according to the estimates of HSP $\hat{\vect{u}}_{t-1}$, and covariance matrix $\widehat{\mathbf{P}}_{\boldsymbol{u}_{t-1}}$ at the previous time instance $t-1$ in a deterministic manner defined by UT (see step 2 in Algorithm 1). The sigma points at time $t-1$ are then dynamically propagated to generate the \textit{priori} prediction of the cardiac electrodynamics $(\vect{u}_{t \mid t-1, i},\vect{v}_{t \mid t-1, i})$ at time $t$ by numerically solving the AP model using finite element method. The statistical prediction $\bar{\vect{u}}_t^-$ is then formed by taking the weighted sample mean of the new ensemble set $\left\{\vect{u}_{t \mid t-1, i}\right\}_{0}^{2 N}$ and the statistical covariance $\mathbf{P}_{\boldsymbol{u}_{t}}^-$ is computed as the weighted sample covariance added by a noise matrix $\boldsymbol{Q}_{\phi}$ which is adopted from Wang et al’s work (see step 3 in Algorithm 1). The \textit{posteriori} correction (i.e., the data assimilation) given the BSPM $\vect{y}_t$ at time $t$ and the noise matrix $\boldsymbol{M}_{\varphi}$ (adopted from Wang et al’s work) is realized by implementing the KF updates (see step 4 in Algorithm 1). The above process is iterated $T$ times, where $T$ is the total number of the discretized time instances.

\begin{algorithm}
%	\color{blue}
	\caption{P-KF algorithm (adapted from Wang \textit{et al.} \cite{wang2009physiological}) }
	\begin{algorithmic} [1]
		  \State Initialization: $\vect{\hat{u}}_0(\vect{s})$
	 
		\State Generate sigma points and the corresponding weights		
		\begin{itemize}
			\item{Sigma points $ \textbf{u}_{t-1,i}(i=0, 2,..., 2N) $}
			\begin{equation}
				\left\{\textbf{u}_{t-1, i}\right\}_{0}^{2N}=\left(\hat{\vect{u}}_{t-1} \quad \hat{\vect{u}}_{t-1} \pm \sqrt{\left(N+\gamma\right) \hat{\mathbf{P}}_{\vect{u}_{t-1}}}\right)
				\nonumber
			\end{equation}
		\end{itemize}

		\begin{itemize}
			\item Weights $W_i$  
			\begin{equation}
				\begin{array}{c}
					W_{0}^{m}=\lambda /\left(N+\gamma\right) \\
					W_{0}^{c}=\lambda /\left(N+\gamma\right)+\left(1-p^{2}+q\right) \\
					W_{i}^{m}=W_{i}^{c}=1 /(2(N+\gamma)) \quad i=1 \ldots 2N
					\nonumber
				\end{array}
			\end{equation}
		\end{itemize}

		\State AP model-based statistical \textit{priori} prediction
		\begin{itemize}
			\item AP model applicaton 
			\begin{equation}
				\left(\textbf{u}_{t \mid t-1, i}, \textbf{v}_{t \mid t-1, i}\right)=\mathcal{F}\left(\textbf{u}_{t-1, i}, \overline{\textbf{v}}_{t-1}\right) \nonumber
			\end{equation}
		\end{itemize}
		
		\begin{itemize}
			\item Statistical Prediction
			
			    \begin{equation*}
			    	\overline{\vect{u}}_{t}^{-}=\sum_{i=0}^{2N} W_{i}^{m} \textbf{u}_{t \mid t-1, i},~~~\overline{\vect{v}}_{t}=\sum_{i=0}^{2N} W_{i}^{m} \textbf{v}_{t \mid t-1, i} 
			    \end{equation*}							
			\vspace{-1em}
			\begin{equation}
				\mathbf{P}_{\boldsymbol{u}_{t}}^{-}=\sum_{i=0}^{2N} W_{i}^{c}\left(\textbf{u}_{t \mid t-1, i}-\overline{\vect{u}}_{t}^{-}\right)\left(\textbf{u}_{t \mid t-1, i}-\overline{\vect{u}}_{t}^{-}\right)^{T}+\mathbf{Q_\phi}
				\nonumber
			\end{equation}
		\end{itemize}

		\State Kalman filter-based \textit{posteriori} correction
		\begin{itemize}
			\item Kalman gain: $\mathbf{K}_{t}=\mathbf{P}_{\vect{u}_{t}}^{-} \mathbf{R}^{T}\left(\mathbf{R P}_{\vect{u}_{t}}^{-} \mathbf{R}^{T}+\mathbf{M_\varphi}\right)^{-1}$
		\end{itemize}
		
		\begin{itemize}
			\item Mean estimation: $\hat{\vect{u}}_{t}=\overline{\vect{u}}_{t}^{-}+\mathbf{K}_{t}\left(\vect{y}_{t}-\mathbf{R} \overline{\vect{u}}_{t}^{-}\right) $
		\end{itemize}

		\begin{itemize}
			\item Covariance update: $	\hat{\mathbf{P}}_{\vect{u}_{t}}=\left(\mathbf{I}-\mathbf{K}_{t} \mathbf{R}\right) \mathbf{P}_{\boldsymbol{u}_{t}}^{-} $
		\end{itemize}
		
		%	\EndFor
	\end{algorithmic}
\end{algorithm}

\subsection{Statistical test for the comparison of $RE$ generated from P-DL and P-KF}
\label{p-value}
We formulate the following hypothesis testing to evaluate if the difference in predictive performance between our P-DL method and the P-KF algorithm with the perfect initial condition (i.e., $ \widehat{\vect{u}}_0(\vect{s}) = \vect{u}_0(\vect{s}) $) is statistically significant:
\begin{align}
H_0: \mu_A&=\mu_B
\\
H_1: \mu_A & \neq \mu_B
\end{align}
where $ \mu_A $and $ \mu_B $ denote the population means of $RE$ yielded from P-DL and P-KF, respectively. $ H_0 $ assumes the $ RE $ from both methods does not have a significant difference, while $ H_1 $ believes otherwise. We engage a two-sample $T$-test to calculate the test statistics by assuming that the calculated $RE$ follows a Gaussian distribution and has an unknown population variance. As such, the test statistics is given as
    \begin{eqnarray}
          T_0^*=\frac{\bar{X}_{A}-\bar{X}_{B}}{\sqrt{S_A^2/n_A+S_B^2/n_B}}
    \end{eqnarray}
where $\bar{X}_{A}=0.1490$ and $\bar{X}_{B}=0.1426$ denote the sample mean of $RE$ yielded from P-DL and P-KF respectively, $S_A=0.0123$ and $S_B=1e-5$ denote the sample standard deviations, and $n_A=n_B=10$ is the repetition number for the experiment. If $H_0$ is true, $T_0^*$ will follows a $t$-distribution with the degrees of freedom given by
  \begin{eqnarray}
  \nu=\frac{(S_A^2/n_A+S_B^2/n_B)^2}{(S_A^2/n_A)^2/(n_A-1)+(S_B^2/n_B)^2/(n_B-1)}
  \end{eqnarray}

The resulted p-value is 0.1343, which is greater than 0.01. This implies the null hypothesis cannot be rejected with a significance level of 0.01. As such, it is legitimate to conclude that there is no significant difference between the performance of our P-DL approach and the P-KF method with the perfect initial condition when solving the inverse ECG problem.
}

\bibliographystyle{IEEEtran}

\bibliography{references}

\begin{IEEEbiography}[{\includegraphics[width=1in,height=1.25in,clip,keepaspectratio]{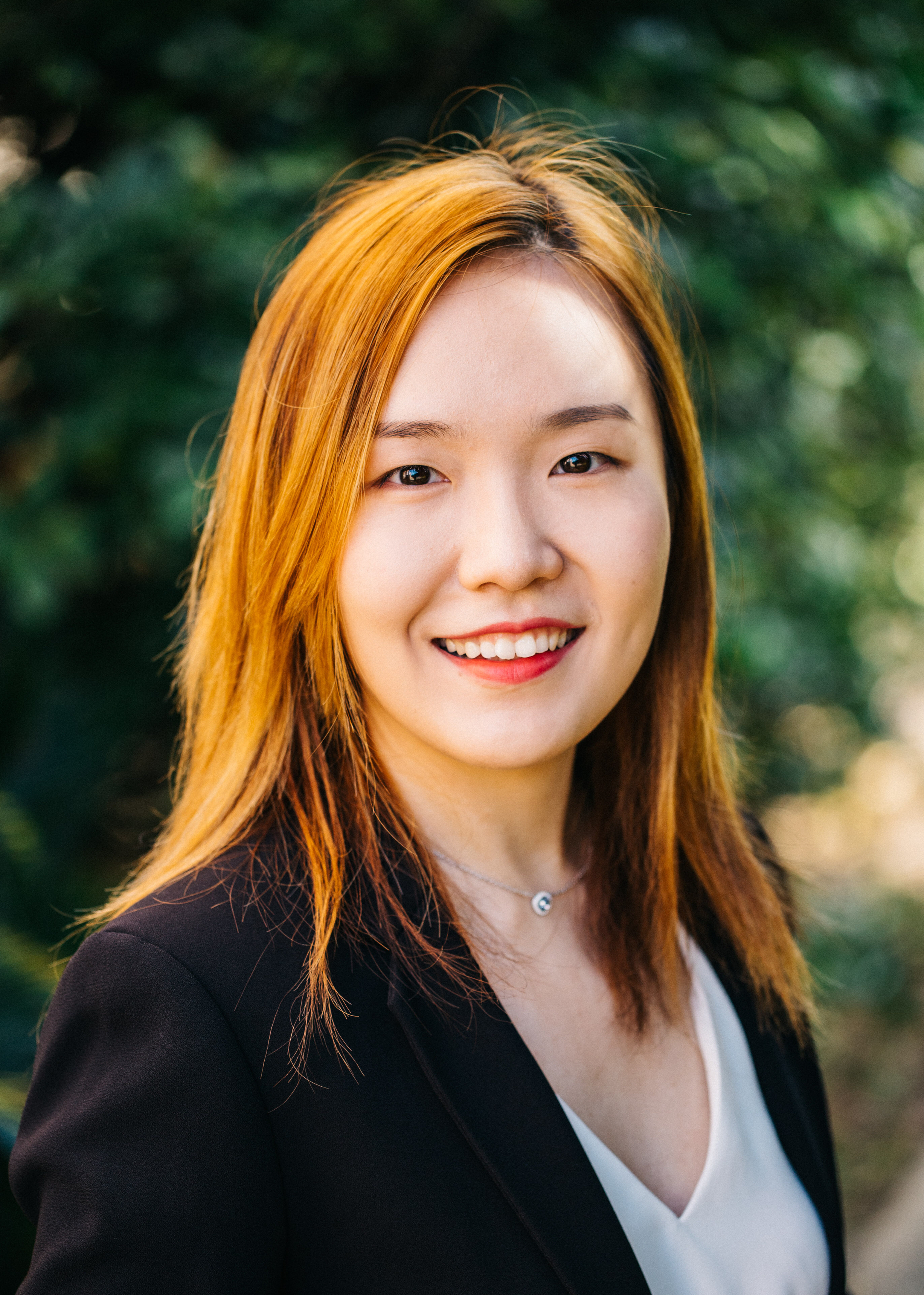}}]%
	{Jianxin Xie} was born in Jiangxi Province, China. She is currently a Ph.D. student in the School of Industrial Engineering and Management with Oklahoma State University. She received her BS degree from Southeast University, China, and MS degree from Florida State University, USA. Her current research interests lie in advanced data analytics, data mining, and physical-statistical modeling with healthcare applications.

\end{IEEEbiography}

\begin{IEEEbiography}[{\includegraphics[width=1in,height=1.25in,clip,keepaspectratio]{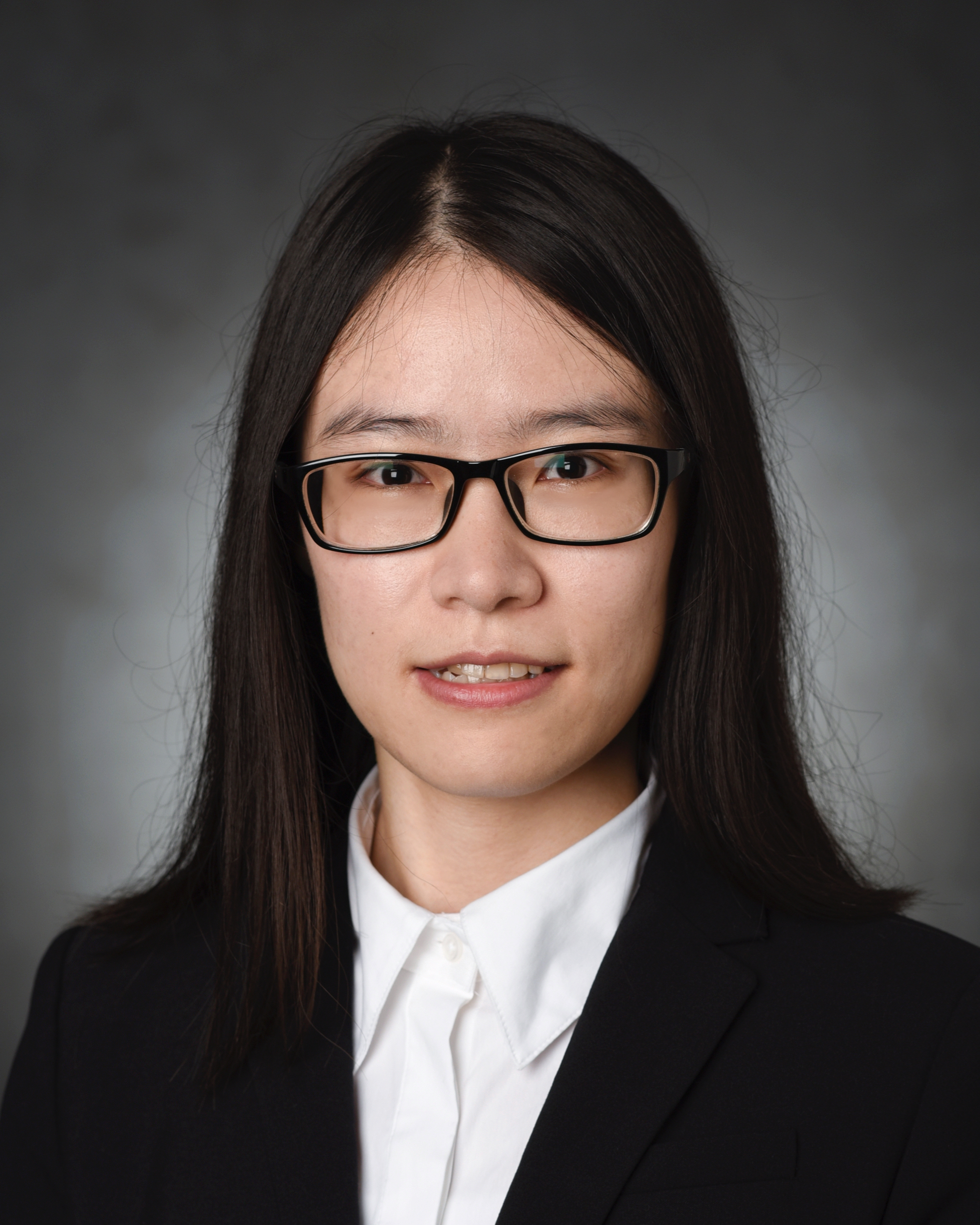}}]%
	{Bing Yao} was born in Nanchong, Sichuan Province, China. She is currently an Assistant Professor in the School of Industrial Engineering and Management with Oklahoma State University. She received her BS degree from University of Science and Technology of China, MS and PhD degrees from the Pennsylvania State University. Her research interests include biomedical and health informatics, computer simulation and optimization, data mining and signal processing, and physical-statistical modeling.
	
\end{IEEEbiography}
% that's all folks
\end{document}